\pdfoutput=1
\documentclass[11pt]{article}

\usepackage[]{EMNLP2023}

\usepackage{times}
\usepackage{latexsym}
\usepackage{tabularx, booktabs}

\usepackage[T1]{fontenc}
\usepackage[utf8]{inputenc}
\usepackage{booktabs}
\usepackage{multirow}

\usepackage{microtype}

\usepackage{inconsolata}

\usepackage{lscape}
\usepackage{subfig}
\usepackage{graphicx}
\usepackage{hyperref}
\usepackage{todonotes}

\title{A Benchmark for Evaluating Machine Translation Metrics\\on Dialects Without Standard Orthography}

\author{Noëmi Aepli$^{1}$ \hspace{0.5cm} Chantal Amrhein$^{1,2}$ \hspace{0.5cm}  Florian Schottmann$^{2,3}$ \hspace{0.5cm}  Rico Sennrich$^{1,4}$ \medskip \\
  $^1$University of Zurich, 
  $^2$Textshuttle,  
  $^3$ETH Zurich,  
  $^4$University of Edinburgh \medskip \\ 
  \texttt{\{naepli,sennrich\}@cl.uzh.ch, \{amrhein,schottmann\}@textshuttle.com}}

\begin{document}
\maketitle

\begin{abstract}
For sensible progress in natural language processing, it is important that we are aware of the limitations of the evaluation metrics we use. In this work, we evaluate how robust metrics are to non-standardized dialects, i.e.\ spelling differences in language varieties that do not have a standard orthography. To investigate this, we collect a dataset of human translations and human judgments for automatic machine translations from English to two Swiss German dialects. We further create a challenge set for dialect variation and benchmark existing metrics' performances. Our results show that existing metrics cannot reliably evaluate Swiss German text generation outputs, especially on segment level. We propose initial design adaptations that increase robustness in the face of non-standardized dialects, although there remains much room for further improvement. 
The dataset, code, and models are available here: \url{https://github.com/textshuttle/dialect_eval}
\end{abstract}

\section{Introduction}

As multilingual NLP models include more and more languages, the community's focus on low-resource languages has also grown. This not only includes languages for which we have ``little data'' but also language varieties and dialects which often pose additional challenges, especially if they do not have a standardized orthography. 
Recent work has shown some progress in classification tasks \citep[e.g.][]{wang-etal-2021-efficient-test, touileb-barnes-2021-interplay, aepli-sennrich-2022-improving} as well as generation tasks where such language varieties appear on the input side only \citep[e.g.][]{zbib-etal-2012-machine,honnet-etal-2018-machine,alam2023codet}. For these scenarios, we can use established evaluation schemes. However, for research towards NLP models \textit{generating} language varieties, \citet{sun-etal-2023-dialect} have shown that current evaluation metrics are not robust to translations into different dialects.

What their evaluation does not consider is that language varieties often lack a standardized orthography and do not adhere to consistent spelling rules. This implies that even \textit{within} a single dialect, notable orthographic variations can be observed, as illustrated in the Swiss German example in Figure \ref{fig:gsw_example}. The same utterance with a similar but different spelling would result in a high word error rate of $\frac{3}{4}$. 

\begin{figure}
\centering
\resizebox{\linewidth}{!}{%
\begin{tabular}{lllll}
\textbf{GSW} &  ... ufere & Webs\colorbox[HTML]{72c5ff}{ii}te & \colorbox[HTML]{72c5ff}{aa}glueg\colorbox[HTML]{72c5ff}{e}t   & w\colorbox[HTML]{72c5ff}{ä}rd\colorbox[HTML]{72c5ff}{e}. \\ % diä   & chönd & äbefalls & 
\textbf{GSW} &  ... ufere     & Webs\colorbox[HTML]{edf8b1}{i}te  & \colorbox[HTML]{edf8b1}{ah}gluegt   & w\colorbox[HTML]{edf8b1}{e}rd\colorbox[HTML]{edf8b1}{ä}. \\ % die   & chönd & ebefalls  & 
\textbf{de}  & ... auf einer & Webseite & angeschaut & werden. \\% diese & können & ebenfalls & 
\textbf{en}  & \multicolumn{4}{c}{\textit{... viewed on a website.}} % These can also be 
\end{tabular}}
\caption{Example sentence that shows the extent of spelling variability in language varieties, here Swiss German dialect (GSW), with German (de) and English (en) translations.}
\label{fig:gsw_example}
\end{figure}

Many languages have multiple regional variants, such as Spanish (Mexican, Argentinean, etc.), French (Canadian, Belgian, etc.), or English (British, American, Australian, Indian, etc.), among others. Such language varieties exhibit various lexical, grammatical, and orthographical distinctions. Importantly, these differences are \textit{standardized}, meaning that they adhere to specific spelling rules and conventions, albeit with variations specific to each variant.
This suggests that if a neural metric is exposed to a sufficient amount of data encompassing various language varieties, it should be able to develop similar representations and provide comparable scores for a given sentence in different varieties. \citet{sun-etal-2023-dialect} show that pre-training a metric on data from multiple dialects indeed makes metrics more inter-dialect robust. 

However, for a substantial number of languages and language varieties, there exists no established standard orthography. Many regions exhibit a dialect continuum where language varieties lack precise boundaries, and each dialect displays a significant range of diversity within itself. Furthermore, when speakers write in their dialect, they follow their individual writing styles. Such kinds of variabilities, as can be observed in the example in Figure \ref{fig:gsw_example}, are much less consistent and localized and will differ significantly between different writers. 
A metric designed to handle these kinds of varieties must be capable of addressing frequent spelling differences, which is considerably more challenging to learn solely from data compared to the standardized language variation differences mentioned in the previous paragraph.

In recent years, embedding-based metrics have gained increasing popularity \citep{sellam-etal-2020-bleurt,rei-etal-2020-comet} which -- in theory -- could be more appropriate for assessing non-standardized language varieties than string-based MT metrics like BLEU \citep{papineni-etal-2002-bleu} or chrF \citep{popovic-2015-chrf}. However, these neural metrics are often not trained on the language varieties in question. Additionally, recent work showed that reference-based learned metrics still rely too much on subword overlap with the reference \citep{hanna-bojar-2021-fine, amrhein-etal-2022-aces}. 

In this work, we follow \citet{sun-etal-2023-dialect} and analyze the dialect robustness of machine translation metrics but specifically focus on non-standardized language varieties that were not seen during pre-training. Our contributions are:
\begin{itemize}
    \item We collect a new dataset and design a challenge set for evaluating MT metrics on two Swiss German dialects.
    \item We benchmark existing string-based and neural metrics on our dataset and find that they are not reliable, especially on segment level.
    \item We propose initial adaptations to make metrics more robust for Swiss German but find that there is still a lot of room for improvement.
\end{itemize} 

\section{Related Work}

There is a substantial amount of research on MT \textit{into} language varieties \citep{scherrer-2011-syntactic,haddow-etal-2013-corpus,fancellu-etal-2014-standard,hassani-2017-kurdish,costa-jussa-etal-2018-neural,lakew-etal-2018-neural,myint-oo-etal-2019-neural,Wan_2020,garcia2022using}. 
Most of these works exclusively evaluate with surface-level metrics like BLEU \citep{papineni-etal-2002-bleu} but some voice their concerns over a lack of reliable evaluation metrics \citep{kumar-etal-2021-machine,bapna-etal-2022-building}.

\citet{sun-etal-2023-dialect} confirm that existing machine translation evaluation metrics are not dialect-robust. They show that it is possible to train more robust metrics by including a language and dialect identification task in a second language model pre-training phase. 
While they focus on inter-dialect robustness between well-defined dialects, i.e.\ Brazilian and Iberian Portuguese, our study focuses on a setting where dialects lack standardized orthography. 
This absence of standardization introduces additional variability, resulting in distinct challenges and necessitating different solutions for MT systems, which need to generalize to often limited data; MT metrics, which need to be robust to spelling differences; and also meta-evaluation, which has its own challenges when collecting human assessments for dialects without standardized orthography as we outline in Section \ref{subsec:human_judgements}.
To investigate how reliable MT metrics are for  non-standardized varieties, we collect a new dataset with human translations and human judgments for MT outputs from English to two Swiss German dialects.

While other works also evaluate MT metrics on language varieties and dialects, \citet{sun-etal-2023-dialect} is closest to our work: \citet{alam2023codet} only look at language varieties on the source side and \citet{riley-etal-2023-frmt} only evaluate language varieties for which a standard was included in the language model pre-training. Both studies also conclude that existing metrics are not robust to dialects. \citet{riley-etal-2023-frmt} further propose a new automated lexical accuracy metric based on term dictionaries, similar to metrics used for automatic speech recognition (ASR) \citep{ali-etal-2017-werd,nigmatulina-etal-2020-asr} which allow for more flexible string matching by using a look-up table of acceptable spellings.
\citeauthor{riley-etal-2023-frmt}'s approach may work well if there is a limited set of term differences between dialects. However, such a metric is difficult to employ for language varieties without standardized spelling rules. Instead, we experiment with increasing dialect robustness by introducing character-level noise during metric training which has been shown to be useful for cross-lingual transfer to language varieties without standardized orthography \citep{aepli-sennrich-2022-improving,srivastava-chiang-2023-fine,blaschke2023does}.

\section{Evaluation Data for Swiss German Dialects}

While we focus on Swiss German because there are enough different MT systems that can be evaluated, Swiss German is by no means the only language where its varieties do not have standardized spelling. Many medium to high-resource languages like Arabic \citep{darwish-etal-2021-panoramic} or Italian \citep{ramponi2022nlp} include dialectal varieties that lack a standardized orthography. 
Additionally, this phenomenon extends to numerous low-resource settings \citep{bird-2022-local}, encompassing a wide array of language varieties across Africa \citep{adebara-abdul-mageed-2022-towards}, Asia \citep{roark-etal-2020-processing,aji-etal-2022-one}, Oceania \citep{solano-etal-2018-development} and the Americas \citep{littell-etal-2018-indigenous, mager-etal-2018-challenges}. Historically, even many language varieties that now have a standardized orthography did not always have one, including English \citep{scragg-1974-history}. This makes our work on robust metrics for non-standardized dialects also relevant for NLP for historical texts.

To measure robustness against non-standardized dialects, we design two new datasets. With the first, we investigate how metrics behave in a realistic setup where we compare them against human judgments.
The second is a challenge set that allows us to investigate score changes between different spellings and compare them to score changes when meaning is changed. This is inspired by similar experiments in \citet{sun-etal-2023-dialect}.

\subsection{Human Judgement Data}
\label{subsec:human_judgements}

In order to realistically evaluate machine translation metrics on Swiss German dialects, it is essential to obtain human-translated reference segments and human judgments for machine-translated translation hypotheses. Since no such data exists for Swiss German, we compile our dataset based on the English NTREX-128 data\footnote{\url{https://github.com/MicrosoftTranslator/NTREX}} \citep{federmann-etal-2022-ntrex}. We selected this dataset because it originates from a standard test set\footnote{\texttt{newstest2019} from the 2019 news translation shared task at WMT \citep{barrault-etal-2019-findings}}, already contains human translations into 128 languages including some regional variants, has a permissive license\footnote{Attribution-ShareAlike 4.0 International (CC BY-SA 4.0)} and offers document context which is important for collecting reliable human judgments \citep{laubli-etal-2018-machine, toral-etal-2018-attaining}. 

\paragraph{Human reference translations:} For the reference translations, we provided two Swiss German translators with the English NTREX-128 source data (i.e.\ 1997 sentences from 123 documents). Translators saw sentences in document context and were asked to translate them into their respective native dialects (i.e.\ Bern and Zurich region). We provided translators with simple instructions where we stated that they must not post-edit machine translation outputs to translate the texts.

\paragraph{Human judgment scores:} The hypotheses come from ten machine translation systems translating from English to Bern dialect and ten systems translating from English to Zurich dialect. For each dialect, we include nine neural MT systems in our rating setup and one rule-based system. 

The neural models are provided by Textshuttle. They are based on a standard Transformer architecture \citep{vaswani2017_attention} trained using different amounts of data, making use of data augmentation techniques like backtranslation \citep{sennrich-etal-2016-improving}. Some of the systems use German as a pivot language. In collaboration with Textshuttle, we decided to evaluate models for which they expect noticeable translation differences and not to compare the nine models that they think would perform the best. 
The rule-based system works by morphosyntactically analyzing the standard German NTREX-128 translation of the English source and then sequentially applying a set of dialect-specific rewriting rules to generate Swiss German output. The system is described in detail in \citet{10.1007/978-3-642-23138-4_9}. The system version used for this task operates word by word without taking syntax into account. Notably, this means that past tense and genitive forms produce unpredictable output because they would require larger changes in the sentence structure. 

We translated the English NTREX-128 source data with each neural system and the German NTREX-128 translation with the rule-based systems and let native dialect speakers rate the outputs via Appraise\footnote{\url{https://github.com/AppraiseDev/Appraise}} \citep{federmann-2018-appraise}, a framework for the evaluation of machine translation outputs. Raters only had access to the source for context because providing the reference could incentivize raters to ``quickly compare the surface forms of translation against reference without understanding'' \citep{freitag-etal-2022-results}. Note that in order to mitigate dialect preference biases as documented by \citet{riley-etal-2023-frmt} and \citet{abu-farha-magdy-2022-effect}, the translators and raters were all native speakers of the dialect they were asked to rate or translate into. We collected continuous Direct Assessment (DA) scores \citep{graham-etal-2013-continuous} where the slider presented to the raters was annotated with Scalar Quality Metric (SQM) labels which increases the rating stability across annotators \citep{kocmi-etal-2022-findings}. Raters viewed segments in a document context and rated translations on the segment level as well as the document level. The document-level ratings are collected to enable future research on document-level metrics; in this study, we only focus on segment-level ratings.

\paragraph{} Ideally, we would recruit professional translators for both the translation and the rating tasks. However, there exist no professional translators for Swiss German. Instead, we recruited translators and annotators from a pool of reliable candidates who already worked on similar Swiss German projects. 
To ensure the quality of the ratings we collect, we included control segments as implemented in Appraise. Based on this control, no raters needed to be excluded.

As Swiss German constitutes a dialect continuum, its various variations lack precise boundaries, and each dialect displays a significant range of diversity within itself. Consequently, during the recruitment process, we placed our trust in the annotators' self-identification of their native dialects. Furthermore, it is worth noting that all our contributors, comprising six women and five men, belong to younger generations, with raters ranging in age from 23 to 30, and translators aged 35 to 40, respectively. This age factor has an impact on their dialect.
All translators and annotators were paid 30 CHF per hour for their work. 

\subsection{Challenge Set}
As an additional evaluation, we compile a challenge set to directly pinpoint how robust metrics are to dialect variability. In the creation of this challenge set, we draw inspiration from the work of \citet{sun-etal-2023-dialect}, who propose measuring inter-dialect robustness by comparing metric scores between two language varieties and between one variety and a version with significant meaning changes. If segment pairs of the latter type are judged more or equally similar by a metric than those of the two varieties, \citet{sun-etal-2023-dialect} argue the metric is not dialect-robust.

We build our challenge set from the collected data presented in the previous section. We filter for all MT hypotheses that humans rated as perfect (i.e.\ received a score of 100). If more than one unique hypothesis exists for a segment, we create all combinations of these hypotheses. For example, if four different machine translation outputs for the same source all receive a perfect human rating, this results in six pairs of semantically equivalent translation hypotheses that feature orthographic differences. For each pair, we then manually create a modified version of one of the hypotheses to change its meaning. Following \citet{sun-etal-2023-dialect}, we consider deletion, insertion, and substitution operations for introducing meaning changes which we randomly assign to each hypothesis pair. All changes are made either to a single word or if necessary a whole phrase. This process results in hypothesis triples as seen in this example:\\

\noindent\resizebox{\linewidth}{!}{%
\begin{tabular}{l}
     \textbf{A:} S\colorbox[HTML]{72c5ff}{e}chs Mitarbeiter s\colorbox[HTML]{72c5ff}{i} wäg\colorbox[HTML]{72c5ff}{e} Verletzige behandlet worde. \\
     \textbf{B:} S\colorbox[HTML]{edf8b1}{ä}chs Mitarbeiter s\colorbox[HTML]{edf8b1}{y} wäg Verletzige behandlet worde. \\ \addlinespace
     \textit{Six members of staff have been treated for injuries.} \\\\\addlinespace 
     \textbf{C:} Sechs Mitarbeiter si wäge Verletzige \textbf{beschtraft} worde.\\ \addlinespace
     \textit{Six members of staff \textbf{were punished because of} injuries.} \vspace{0.35cm} 
\end{tabular}}

Hypotheses A and B are semantically equivalent but exhibit spelling differences. Hypothesis C is 
very similar to hypothesis A on the surface level but differs significantly in meaning. During evaluation, metrics will have access to one of these hypotheses, as well as the reference and/or the source (depending on whether it is a reference-free or reference-based metric). We describe how we compare the different scores for these hypotheses in Section \ref{sec:eval}.

\section{Experiment Setup}

\subsection{Benchmarking Existing Metrics}

To document the performance of current MT metrics on dialects without a standard orthography, we evaluate the following metrics:

\begin{itemize}
    \item \textbf{BLEU}\footnote{computed with \href{https: //github.com/mjpost/sacrebleu/}{SacreBLEU} \cite{post-2018-call}, signature: nrefs:1|case:mixed|eff:no|tok:13a|smooth:exp|version:2.3.0.} \citep{papineni-etal-2002-bleu}, a string-based metric with a brevity penalty that calculates the word-level n-gram precision between a translation and one or multiple references.
    \item \textbf{chrF++}\footnote{computed with \href{https: //github.com/mjpost/sacrebleu/}{SacreBLEU} \cite{post-2018-call}, signature: nrefs:1|case:mixed|eff:yes|nc:6|nw:0|space:no|version:2.3.0.} \citep{popovic-2017-chrf}, another string-based metric that provides a character n-gram, word unigram, and bigram F-score by computing overlaps between the hypothesis and reference translation.
\end{itemize}

We expect surface-level, string-based metrics to perform badly on dialects without standard spelling rules as they are entirely based on overlap with a reference translation. These are also the metrics used by most works that explored text generation for language varieties without standardized orthography \citep[e.g.][]{jeblee-etal-2014-domain,meftouh-etal-2015-machine,kumar-etal-2021-machine}. We further benchmark the following neural metrics:

\begin{itemize}
    \item \textbf{COMET-20}\footnote{\href{https://huggingface.co/Unbabel/wmt20-comet-da}{wmt20-comet-da}} \citep{rei-etal-2020-unbabels} and \textbf{COMET-22}\footnote{\href{https://huggingface.co/Unbabel/wmt22-comet-da}{wmt22-comet-da}} \citep{rei-etal-2022-comet}, two reference-based neural metrics built on the COMET framework \citep{rei-etal-2020-comet}. These are trained neural metrics that are built on top of a large, pre-trained language model and are fine-tuned on human judgment data from previous metric evaluation campaigns. COMET-20 is fine-tuned to predict DA scores. COMET-22 is an ensemble between a COMET-20-like model and a multi-task model that predicts segment-level Multidimensional Quality Metric (MQM) scores \citep{uszkoreit-lommel-2013-multidimensional} as well as word-level error tags.
    \item \textbf{COMET-20-QE}\footnote{\href{https://huggingface.co/Unbabel/wmt20-comet-qe-da}{wmt20-comet-qe-da}} \citep{rei-etal-2020-unbabels} and \textbf{COMET-Kiwi}\footnote{\href{https://huggingface.co/Unbabel/wmt22-cometkiwi-da}{wmt22-cometkiwi-da}} \citep{rei-etal-2022-comet}, two reference-free neural metrics for quality estimation. COMET-20-QE is trained similarly to COMET-20 and COMET-KIWI to COMET-22, but both versions do not have access to the reference during training on human judgments.
\end{itemize}

While these metrics go beyond surface-level comparisons to the reference due to their hidden representations and embedding-based nature, we expect that they still struggle to reliably evaluate translations into Swiss German for several reasons: First, no Swiss German data was included for pre-training the language model \citep[XLM-R;][]{DBLP:journals/corr/abs-1911-02116} that is used as the basis for training COMET. Second, neural metrics are often fine-tuned on Standard German data which shares many similar words with Swiss German and could falsely bias metrics towards Standard German spelling. Third, reference-based metrics have been shown to still be influenced by surface overlap with the reference \citep{hanna-bojar-2021-fine, amrhein-etal-2022-aces} which is a disadvantage in situations where numerous spelling variations exist.

\subsection{Developing Dialect-Robust Metrics}\label{sec:setup}

Similar to \citet{sun-etal-2023-dialect}, we also experiment with training more robust metrics but we focus on robustness against non-standardized dialects rather than inter-dialect robustness. The following list summarizes our metrics:

\begin{itemize}
    \item \textbf{COMET-REF} and \textbf{COMET-QE}, a baseline trained as a reference to compare our modifications to because our COMET models differ slightly from COMET-20 and COMET-22 (see details below).
    \item \textbf{+gsw}, same as the baseline but the pre-trained model is fine-tuned on Swiss German data before the COMET models are fine-tuned on human judgment data. This is similar to the second pre-training phase for the inter-dialect-robust metric proposed in \citet{sun-etal-2023-dialect}. However, we do not include the additional language and dialect identification task during continued pre-training as we do not have dialect labels for the Swiss German pre-training data.
    \item \textbf{+noise}, same as the baseline but during the fine-tuning process on human judgment data we introduce character-level noise. 
    This is inspired by previous work that showed that this method allows for better cross-lingual transfer to closely related languages \citep{aepli-sennrich-2022-improving,srivastava-chiang-2023-fine}. \citet{blaschke2023does} hypothesize that injecting noise into standard language data results in a similar tokenization rate as for unseen dialects. 
    We apply noise injection to all languages within the COMET fine-tuning dataset that have an alphabetic writing system, therefore excluding languages like Chinese which were not considered in the original work introducing character-level noise. Following \citet{aepli-sennrich-2022-improving}, we inject character-level noise (essentially typos) into a random selection of 15\% of the tokens within each sentence. Specifically, we alter, delete, or add one character per chosen token. We execute this process using the characters specific to the relevant language, taking into account all characters that occur more than 1,000 times in the respective dataset. We apply this noise injection to all segments, including the source, translation, and reference segments.
\end{itemize}

We provide details of how we trained those models here:

\paragraph{Continued pre-training of XLM-R}  

To expose our models to Swiss German data, we modify the encoder model upon which COMET models are usually based: XLM-RoBERTa\footnote{\href{https://huggingface.co/xlm-roberta-base}{xlm-roberta-base}} \citep{DBLP:journals/corr/abs-1911-02116}. 
We continue the training of the XLM-R model on SwissCrawl\footnote{\href{https://icosys.ch/swisscrawl}{swisscrawl}} \citep{linder2020crawler}, a corpus containing 500K dialect sentences crawled from the web in late 2019.
For the continued pre-training, we work with the Huggingface Transformers library\footnote{\url{https://github.com/huggingface/transformers}} \citep{wolf-etal-2020-transformers}, following the default configurations for language model fine-tuning which involves a training duration of three epochs.

\paragraph{Training COMET models}

We train COMET models using the official code base\footnote{\url{https://github.com/Unbabel/COMET}} with the default settings from version 2.0.2. We use the ``regression model'' configuration for the reference-based models and the ``referenceless model'' configuration for the reference-free models. Our models are trained on the direct assessment data collected by the organizers of the WMT news translation task spanning the years 2017 to 2021 (2021 as dev set)\footnote{\url{https://github.com/Unbabel/COMET/tree/master/data}} \citep{bojar-etal-2017-findings, bojar-etal-2018-findings, barrault-etal-2019-findings, barrault-etal-2020-findings, akhbardeh-etal-2021-findings}. It is important to highlight that our models are not directly comparable to the original WMT shared task COMET models, for which the 2020 models were exclusively trained on data from 2017-2019 and the 2022 models used a different configuration.

\subsection{Evaluation}
\label{sec:eval}
We evaluate our metrics in five different ways. 
For the human judgment data, we compute two scores on system (sys) and two on segment (seg) level using the reference implementation from the WMT metrics shared task\footnote{\url{https://github.com/google-research/mt-metrics-eval}} \citep{freitag-etal-2022-results}, except for \textit{success rate} where we use our own implementation.

\paragraph{System level} The \textit{pairwise accuracy} as defined by \citet{kocmi-etal-2021-ship}, measures the accuracy with which a metric agrees with human preference between pairs of systems where the human ratings are significantly different according to a two-sided Wilcoxon test. 
Note that the score difference between the two systems is not important in this analysis. 
Furthermore, we provide results for the \textit{sys-level Pearson correlation}, quantifying the strength of the linear relationship between metrics and human judgment scores for systems.

\paragraph{Segment level} At the segment level, our evaluation includes the \textit{seg-level accuracy} with an optimized tie threshold, which resembles a global accuracy but also acknowledges metrics for correctly predicting tied human judgment scores \citep{deutsch2023ties}. Further, we present the \textit{seg-level Kendall correlation}, akin to pairwise accuracy but employing a distinct normalization technique.

\paragraph{Challenge set} For the challenge set, we compute the \textit{success rate} (seg level) following \citet{sun-etal-2023-dialect}. 
This measures the accuracy with which a metric assigns more similar scores (s) to two equivalent translations A and B compared to a version with a semantic change C. Consequently, a metric is considered robust to non-standardized dialects for a segment if the score difference between $s_A$ and $s_B$ is smaller than the score difference between $s_C$ and either $s_A$ or $s_B$ (depending on which score is smaller):

\begin{equation}
    |s_A - s_B| < min(s_A, s_B) - s_C
\end{equation}

\section{Results}
\label{sec:results}

\begin{table*}[]
\centering
\begin{tabularx}{\textwidth}{@{\extracolsep{3pt}}rccccccccc}
 &
  \multicolumn{3}{c}{system-level} &
  \multicolumn{6}{c}{segment-level} \\ \cmidrule(lr){2-4} \cmidrule(lr){5-10} \addlinespace
 &
  \multicolumn{1}{c}{\bf{pairwise}} &
  \multicolumn{2}{c}{\bf{Pearson}} &
  \multicolumn{2}{c}{\bf{tie-optim.}} &
  \multicolumn{2}{c}{\bf{Kendall}} &
  \multicolumn{2}{c}{\bf{success}} \\
 &
  \multicolumn{1}{c}{\bf{accuracy}} &
  \multicolumn{2}{c}{\bf{correlation}} &
  \multicolumn{2}{c}{\bf{accuracy}} &
  \multicolumn{2}{c}{\bf{correlation}} &
  \multicolumn{2}{c}{\bf{rate}} \\
 &
  \multicolumn{1}{l}{} &
  \multicolumn{1}{c}{BE} &
  \multicolumn{1}{c}{ZH} &
  \multicolumn{1}{c}{BE} &
  \multicolumn{1}{c}{ZH} &
  \multicolumn{1}{c}{BE} &
  \multicolumn{1}{c}{ZH} &
  \multicolumn{1}{c}{BE} &
  \multicolumn{1}{c}{ZH} \\
BLEU &
  \cellcolor[HTML]{7EC0D0}0.740 &
  \cellcolor[HTML]{359CE3}0.728 &
  \cellcolor[HTML]{359CE3}0.587 &
  \cellcolor[HTML]{8AC6CC}0.544 &
  \cellcolor[HTML]{9ACEC8}0.560 &
  \cellcolor[HTML]{6EB8D4}0.142 &
  \cellcolor[HTML]{75BCD2}0.163 &
  \cellcolor[HTML]{41A2E0}0.135 &
  \cellcolor[HTML]{61B2D7}0.194 \\
chrF &
  \cellcolor[HTML]{87C5CD}0.753 &
  \cellcolor[HTML]{6EB8D4}0.806 &
  \cellcolor[HTML]{58ADDA}0.665 &
  \cellcolor[HTML]{359CE3}0.486 &
  \cellcolor[HTML]{359CE3}0.478 &
  \cellcolor[HTML]{359CE3}0.076 &
  \cellcolor[HTML]{359CE3}0.079 &
  \cellcolor[HTML]{359CE3}0.121 &
  \cellcolor[HTML]{359CE3}0.145 \\
COMET-20 &
  \cellcolor[HTML]{91CACA}0.766 &
  \cellcolor[HTML]{8EC8CB}0.849 &
  \cellcolor[HTML]{9ED0C7}0.816 &
  \cellcolor[HTML]{A9D6C4}0.565 &
  \cellcolor[HTML]{B6DCC0}0.583 &
  \cellcolor[HTML]{A6D4C5}0.205 &
  \cellcolor[HTML]{A7D5C4}0.227 &
  \cellcolor[HTML]{ACD7C3}0.250 &
  \cellcolor[HTML]{C0E1BE}0.298 \\
COMET-22 &
  \cellcolor[HTML]{91CACA}0.766 &
  \cellcolor[HTML]{B1DAC2}0.897 &
  \cellcolor[HTML]{C5E4BC}0.901 &
  \cellcolor[HTML]{B0D9C2}0.570 &
  \cellcolor[HTML]{BBDFBF}0.587 &
  \cellcolor[HTML]{93CBCA}0.184 &
  \cellcolor[HTML]{9BCFC8}0.212 &
  \cellcolor[HTML]{A5D4C5}0.243 &
  \cellcolor[HTML]{C7E5BC}0.306 \\
COMET-20-QE &
  \cellcolor[HTML]{50A9DC}0.675 &
  \cellcolor[HTML]{A1D2C6}0.875 &
  \cellcolor[HTML]{B8DDC0}0.872 &
  \cellcolor[HTML]{55ACDB}0.508 &
  \cellcolor[HTML]{63B3D7}0.516 &
  \cellcolor[HTML]{67B5D6}0.134 &
  \cellcolor[HTML]{5FB1D8}0.134 &
  \cellcolor[HTML]{3EA0E1}0.131 &
  \cellcolor[HTML]{43A3E0}0.161 \\
COMET-KIWI &
  \cellcolor[HTML]{359CE3}0.636 &
  \cellcolor[HTML]{D9EEB7}0.952 &
  \cellcolor[HTML]{B9DEBF}0.876 &
  \cellcolor[HTML]{7EC0CF}0.536 &
  \cellcolor[HTML]{78BDD1}0.533 &
  \cellcolor[HTML]{72BAD3}0.146 &
  \cellcolor[HTML]{65B4D6}0.142 &
  \cellcolor[HTML]{A3D3C6}0.240 &
  \cellcolor[HTML]{B9DEC0}0.290 \\ \addlinespace \hline \addlinespace
COMET-REF &
  \cellcolor[HTML]{7EC0D0}0.740 &
  \cellcolor[HTML]{99CEC8}0.864 &
  \cellcolor[HTML]{93CBCA}0.793 &
  \cellcolor[HTML]{ACD7C3}0.567 &
  \cellcolor[HTML]{A6D4C5}0.570 &
  \cellcolor[HTML]{90C9CB}0.180 &
  \cellcolor[HTML]{8DC8CB}0.194 &
  \cellcolor[HTML]{91CACA}0.221 &
  \cellcolor[HTML]{86C4CD}0.234 \\
+ gsw &
  \cellcolor[HTML]{A3D3C5}0.792 &
  \cellcolor[HTML]{B8DDC0}\textbf{0.906} &
  \cellcolor[HTML]{B3DBC1}0.862 &
  \cellcolor[HTML]{EDF8B1}0.611 &
  \cellcolor[HTML]{EDF8B1}0.627 &
  \cellcolor[HTML]{EDF8B1}\textbf{0.286} &
  \cellcolor[HTML]{EDF8B1}\textbf{0.317} &
  \cellcolor[HTML]{EDF8B1}0.320 &
  \cellcolor[HTML]{EDF8B1}0.347 \\
+ noise &
  \cellcolor[HTML]{75BCD2}0.727 &
  \cellcolor[HTML]{D1EAB9}\textbf{0.940} &
  \cellcolor[HTML]{C6E4BC}\textbf{0.903} &
  \cellcolor[HTML]{A3D3C5}0.561 &
  \cellcolor[HTML]{A2D2C6}0.567 &
  \cellcolor[HTML]{B5DCC0}\textbf{0.223} &
  \cellcolor[HTML]{ACD7C3}\textbf{0.233} &
  \cellcolor[HTML]{A0D1C6}0.237 &
  \cellcolor[HTML]{B9DEC0}0.290 \\
+ gsw + noise &
  \cellcolor[HTML]{A3D3C5}0.792 &
  \cellcolor[HTML]{C0E1BE}\textbf{0.917} &
  \cellcolor[HTML]{B6DCC0}\textbf{0.868} &
  \cellcolor[HTML]{D8EDB7}0.597 &
  \cellcolor[HTML]{E5F4B4}0.621 &
  \cellcolor[HTML]{DFF1B5}\textbf{0.271} &
  \cellcolor[HTML]{E2F2B4}\textbf{0.304} &
  \cellcolor[HTML]{CEE8BA}0.287 &
  \cellcolor[HTML]{D7EDB7}0.323 \\ \addlinespace
COMET-QE-KIWI &
  \cellcolor[HTML]{359CE3}0.636 &
  \cellcolor[HTML]{5CAFD9}0.781 &
  \cellcolor[HTML]{63B3D7}0.689 &
  \cellcolor[HTML]{359CE3}0.486 &
  \cellcolor[HTML]{58ADDA}0.507 &
  \cellcolor[HTML]{4DA8DD}0.104 &
  \cellcolor[HTML]{44A3DF}0.099 &
  \cellcolor[HTML]{3A9EE2}0.127 &
  \cellcolor[HTML]{359CE3}0.145 \\
+ gsw &
  \cellcolor[HTML]{C8E5BC}0.844 &
  \cellcolor[HTML]{EDF8B1}\textbf{0.978} &
  \cellcolor[HTML]{EDF8B1}\textbf{0.987} &
  \cellcolor[HTML]{D5ECB8}0.595 &
  \cellcolor[HTML]{BBDFBF}0.587 &
  \cellcolor[HTML]{D3EBB8}\textbf{0.257} &
  \cellcolor[HTML]{D2EAB9}\textbf{0.283} &
  \cellcolor[HTML]{D3EBB9}0.292 &
  \cellcolor[HTML]{C0E1BE}0.298 \\
+ noise &
  \cellcolor[HTML]{50A9DC}0.675 &
  \cellcolor[HTML]{BEE0BE}\textbf{0.915} &
  \cellcolor[HTML]{9ED0C7}\textbf{0.817} &
  \cellcolor[HTML]{6CB7D4}0.524 &
  \cellcolor[HTML]{72BAD3}0.528 &
  \cellcolor[HTML]{79BED1}\textbf{0.154} &
  \cellcolor[HTML]{72BAD3}\textbf{0.158} &
  \cellcolor[HTML]{4EA8DC}0.149 &
  \cellcolor[HTML]{52AADC}0.177 \\
+ gsw + noise &
  \cellcolor[HTML]{EDF8B1}\textbf{0.896} &
  \cellcolor[HTML]{E5F4B3}\textbf{0.968} &
  \cellcolor[HTML]{EAF6B2}\textbf{0.981} &
  \cellcolor[HTML]{C2E2BD}0.582 &
  \cellcolor[HTML]{C6E4BC}0.596 &
  \cellcolor[HTML]{C9E6BB}\textbf{0.246} &
  \cellcolor[HTML]{C7E5BC}\textbf{0.269} &
  \cellcolor[HTML]{C1E2BD}0.273 &
  \cellcolor[HTML]{AAD6C4}0.274
\end{tabularx}
\caption{Results for the baselines metrics (above) and our trained metrics (below) on system level (left) and segment level (right). Darker shades indicate lower scores.
Bold denotes statistically significant improvement compared to their respective baselines COMET-REF or COMET-QE-KIWI. There is no information about significance for tie-optim. accuracy (columns 4-5) and success rate (columns 8-9).
Note that BE and ZH represent the abbreviations for the two Swiss German (GSW) dialect regions under consideration.}
\label{tab:all_results}
\end{table*}

Table \ref{tab:all_results} provides a comprehensive summary of our results with scores for existing metrics (top), COMET models trained for this work (bottom), system-level evaluations (left), and segment-level evaluations (right). 
Additional results can be found in the appendices. Appendix \ref{app:continued_pretrain} contains results related to the incorporation of additional languages in the pre-training process, Appendix \ref{app:wmt_correlations} presents an evaluation of performance on an official WMT benchmark, and Appendix \ref{app:plots} presents pairwise accuracy plots for our metrics.

\paragraph{Existing vs.\ GSW metrics} As expected, the surface-level metrics perform worse than trained metrics in almost all evaluations. Our baseline metrics often perform a bit worse than the existing COMET metrics, this is particularly true for our reference-free model. However, continued pre-training on Swiss German data improves their performance considerably and they strongly outperform existing metrics. This highlights the importance of the model to have seen the target language (variety) during the language model pre-training. It also shows that metrics can be extended to include new languages and language varieties with limited effort although this impacts their performance on other language pairs as we show in Appendix \ref{app:wmt_correlations}. Continued pre-training on multiple languages and language varieties can mitigate this effect (see Appendix \ref{app:continued_pretrain}).

\paragraph{Noise injection} While continued LM pre-training on Swiss German data generally outperforms noise injection during task fine-tuning, we still see gains over the baselines. This suggests that metrics that were trained on noised data are more robust to unseen language (varieties) and may be a good strategy for language (varieties) without sufficient data for continued pre-training. Combining both continued pre-training and noise injection generally does not lead to further improvements. 

\paragraph{Reference-based vs reference-free} 
While both types of metrics perform similarly with continued pre-training on Swiss German, both existing reference-free metrics perform worse than the existing reference-based metrics in the segment-level evaluations. Since these metrics did not see any Swiss German during the pre-training phase, having access to the reference as an anchor might help the reference-based metrics for unseen languages. \citet{amrhein-etal-2022-aces} reported a similar finding where the reference acted as an anchor when metrics were used to identify copied source sentences.

\paragraph{Challenge set} The success rate for all metrics is extremely low. Metrics assign more similar scores to a hypothesis with a semantic change than to a different translation hypothesis in the majority of cases. Again, continued pre-training on Swiss German results in the best metric performance. However, even these scores are lower than a random success rate of 50\% by far. Our findings highlight that even though system-level correlations may seem convincing, none of the metrics studied in this work are robust to non-standardized dialect variations. \\

Since our results show that there is still significant room for improvement toward metric robustness to non-standardized language varieties, we provide suggestions for future work.

\section{Open Questions}
We hope that our benchmark inspires more work on robust evaluation metrics for language varieties in the future. In this section, we list several directions we think are worthwhile exploring:

\paragraph{Expanding the benchmark:} We were not able to include additional language varieties in our benchmark at the time because we could not find enough \textit{different} machine translation systems that translate into these varieties. While we recognize that without reliable metrics this is a ``chicken-and-egg'' problem, we still advocate for more MT research that focuses on translating \textit{into} language varieties. Expanding our benchmark would not only allow us to draw more general conclusions but would also help with sample size for the \textit{pairwise accuracy} analysis \citep{kocmi-etal-2021-ship} since we find that a large number of systems are required for confident results.

\paragraph{More focus on segment level:} Segment-level metric scores tend to be much less correlated with human judgments when contrasted with system-level correlations \cite{freitag-etal-2022-results} and have also been shown to be unreliable in downstream tasks \citep{moghe-etal-2023-extrinsic}. We hope that future work aimed at enhancing metric performance on our challenge set will also contribute to greater metric reliability on segment level in general, as over-reliance on reference overlap is also a problem for languages with standardized spelling \citep{hanna-bojar-2021-fine, amrhein-etal-2022-aces}.

\paragraph{Training neural metrics that model character-level similarities:} A segment in a dialect often resembles a reference in certain characters only rather than in full words (see Figure \ref{fig:gsw_example} as an example). As the underlying language models of neural evaluation metrics use a fixed tokenization scheme that was learned on text that likely does not include many examples of language varieties, these similarities might be hard to account for by the neural metric. Thus, we believe that character-based language models, such as Canine \citep{clark-etal-2022-canine}, could provide a better basis for neural evaluation metrics to model character-level similarities. 

\section{Conclusion}
We evaluated the reliability of machine translation metrics when evaluating dialects without standard orthographies. As part of this work, we collected a new dataset consisting of human translations, human judgments, and a challenge set from English to two Swiss German dialects. We benchmark several existing metrics and find that they are not robust to variation featured by non-standardized dialects. Based on this finding, we explore several modifications that allow us to train metrics that are more robust towards spelling variation. Our results show that there is still a lot of room for improvement and we offer a set of recommendations for future work on dialect robust metrics.

\section*{Limitations}

The goal of this work is to evaluate and develop machine translation metrics that take into account the spelling variability of dialects and languages without established writing norms. We recognize that evaluating metrics on varieties from different languages would help generalize our results. However, we were not able to find enough differing machine translation systems that translate \textit{into} the same language variety for other languages. Therefore, we had to limit this study to two Swiss German dialects. We hope to include further language varieties in our benchmark in the future (when such machine translation systems become available) to encourage research toward metrics that are reliable for many non-standardized language varieties.

We did our best to avoid dialectal preference bias within our annotators by selecting only annotators who consider themselves native speakers of the respective dialect. However, as Swiss German is a dialect continuum, this can only be controlled to a certain degree.

\section*{Ethics Statement}

This work includes the compilation of a new dataset as a test set for evaluating various machine translation metrics. All translators and annotators were compensated at a rate of 30 CHF per hour. Our dataset is based on a publicly available dataset and will be released under the same license for future use. \textbf{Intended use:} The dataset and the models resulting from this work are intended to be used by the research community to evaluate machine translation metrics.

\section*{Acknowledgements}
We thank Yves Scherrer for providing the rule based systems and helpful comments. Furthermore, we thank Annette Rios, Tom Kocmi, Mathias Müller, and the anonymous reviewers for their valuable inputs. We are also grateful to the Swiss German raters and translators for their important contribution. This work was supported by the Swiss National Science Foundation (project nos. 191934 \& 176727), Textshuttle, and the Department of Computational Linguistics at the University of Zurich.

\bibliography{anthology,custom}
\bibliographystyle{acl_natbib}

\newpage
\appendix

\section{Appendix}
\label{sec:appendix}

\subsection{Mixed Continued Pre-training}\label{sec:app-mix}
\label{app:continued_pretrain}
In our main experiments in Section \ref{sec:results}, we evaluate continued language model pre-training only on Swiss German data. While this increases the performance on our benchmark, it remains unclear whether this leads to a ``specialized'' metric that does not perform well on other language pairs. 
We will evaluate this in the next section, but first, we introduce a set of contrastive models that are less specialized to Swiss German. 
Continued pre-training for contrastive models involves incorporating mixed data from five languages apart from Swiss German, namely: German (de), English (en), French (fr), Hindi (hi), and Chinese (zh). We train one metric based on XLM-R with continued pre-training only on these five languages (``5 langs''), and another one where we also add GSW to the training data (``6 langs''). For both settings, we also test character-level noise in the COMET fine-tuning step, as described in Section \ref{sec:setup}.
The data for the five additional languages is sourced from the CC-100 corpus\footnote{\url{https://data.statmt.org/cc-100/}} \citep{wenzek-etal-2020-ccnet}, which is a reconstructed version of XLM-R's training dataset. Specifically, we utilize the first 100,000 sentences from the training data of each language.

Table \ref{tab:mixed} shows the results we obtained from incorporating mixed data into the continued LM pre-training. We see a similar effect as when continuing the pre-training only on GSW in the main results in Section \ref{sec:results}. The performance of the metrics increases in all evaluations. Comparing these results to the metric where we only continued pre-training on Swiss German (+6 langs vs. +gsw ), the results are comparable and often not significantly different. In the next section, we investigate how these metrics behave on other language pairs.

\noindent
\begin{table*}[]
%\begin{center}
\begin{tabularx}{\textwidth}{@{\extracolsep{3pt}}rccccccccc}
 &
  \multicolumn{3}{c}{system-level} &
  \multicolumn{6}{c}{segment-level} \\ \cmidrule(lr){2-4} \cmidrule(lr){5-10} \addlinespace
 &
  \multicolumn{1}{c}{\bf{pairwise}} &
  \multicolumn{2}{c}{\bf{Pearson}} &
  \multicolumn{2}{c}{\bf{tie-optim.}} &
  \multicolumn{2}{c}{\bf{Kendall}} &
  \multicolumn{2}{c}{\bf{success}} \\
 &
  \multicolumn{1}{c}{\bf{accuracy}} &
  \multicolumn{2}{c}{\bf{correlation}} &
  \multicolumn{2}{c}{\bf{accuracy}} &
  \multicolumn{2}{c}{\bf{correlation}} &
  \multicolumn{2}{c}{\bf{rate}} \\
&
  \multicolumn{1}{l}{} &
  \multicolumn{1}{c}{BE} &
  \multicolumn{1}{c}{ZH} &
  \multicolumn{1}{c}{BE} &
  \multicolumn{1}{c}{ZH} &
  \multicolumn{1}{c}{BE} &
  \multicolumn{1}{c}{ZH} &
  \multicolumn{1}{c}{BE} &
  \multicolumn{1}{c}{ZH} \\
COMET-REF &
  0.740 &
  0.864 &
  0.793 &
  0.567 &
  0.570 &
  0.180 &
  0.194 &
  0.221 &
  0.234 \\
+ noise &
  0.727 &
  \textbf{0.940} &
  \textbf{0.903} &
  0.561 &
  0.567 &
  \textbf{0.223} &
  \textbf{0.233} &
  0.237 &
  0.290 \\
+ gsw &
  0.792 &
  \textbf{0.906} &
  0.862 &
  0.611 &
  0.627 &
  \textbf{0.286} &
  \textbf{0.317} &
  0.320 &
  0.347 \\
+ gsw + noise &
  0.792 &
  \textbf{0.917} &
  \textbf{0.868} &
  0.597 &
  0.621 &
  \textbf{0.271} &
  \textbf{0.304} &
  0.287 &
  0.323 \\
+ 5 langs &
  \cellcolor[HTML]{99CEC8}0.766 &
  \cellcolor[HTML]{95CCC9}0.877 &
  \cellcolor[HTML]{69B6D5}0.774 &
  \cellcolor[HTML]{A3D3C5}0.561 &
  \cellcolor[HTML]{AAD6C4}0.583 &
  \cellcolor[HTML]{A2D2C6}\textbf{0.212} &
  \cellcolor[HTML]{A3D3C5}\textbf{0.230} &
  \cellcolor[HTML]{8BC7CC}0.235 &
  \cellcolor[HTML]{82C2CE}0.274 \\
+ 5 langs + noise &
  \cellcolor[HTML]{99CEC8}0.766 &
  \cellcolor[HTML]{C6E4BC}\textbf{0.938} &
  \cellcolor[HTML]{B1DAC2}\textbf{0.890} &
  \cellcolor[HTML]{B0D9C2}0.570 &
  \cellcolor[HTML]{B9DEBF}0.593 &
  \cellcolor[HTML]{BFE1BE}\textbf{0.241} &
  \cellcolor[HTML]{B9DEBF}\textbf{0.256} &
  \cellcolor[HTML]{A3D3C5}0.265 &
  \cellcolor[HTML]{8BC7CC}0.290 \\
+ 6 langs &
  \cellcolor[HTML]{B2DAC1}0.805 &
  \cellcolor[HTML]{C2E2BD}\textbf{0.932} &
  \cellcolor[HTML]{AFD9C2}\textbf{0.887} &
  \cellcolor[HTML]{D1EAB9}0.592 &
  \cellcolor[HTML]{DCEFB6}0.616 &
  \cellcolor[HTML]{EDF8B1}\textbf{0.286} &
  \cellcolor[HTML]{ECF7B2}\textbf{0.316} &
  \cellcolor[HTML]{EDF8B1}0.357 &
  \cellcolor[HTML]{EDF8B1}0.452 \\
+ 6 langs + noise &
  \cellcolor[HTML]{A1D2C6}0.779 &
  \cellcolor[HTML]{D5ECB8}\textbf{0.956} &
  \cellcolor[HTML]{C1E2BD}\textbf{0.917} &
  \cellcolor[HTML]{DBEFB6}0.599 &
  \cellcolor[HTML]{E5F4B4}0.622 &
  \cellcolor[HTML]{E8F5B3}\textbf{0.282} &
  \cellcolor[HTML]{E7F5B3}\textbf{0.311} &
  \cellcolor[HTML]{D1EAB9}0.323 &
  \cellcolor[HTML]{C1E2BD}0.379 \\ \addlinespace
COMET-QE-KIWI &
  0.636 &
  0.781 &
  0.689 &
  0.486 &
  0.507 &
  0.104 &
  0.099 &
  0.127 &
  0.145 \\
+ noise &
  0.675 &
  \textbf{0.915} &
  \textbf{0.817} &
  0.524 &
  0.528 &
  \textbf{0.154} &
  \textbf{0.158} &
  0.149 &
  0.177 \\
+ gsw &
  0.844 &
  \textbf{0.978} &
  \textbf{0.987} &
  0.595 &
  0.587 &
  \textbf{0.257} &
  \textbf{0.283} &
  0.292 &
  0.298 \\
+ gsw + noise &
  \textbf{0.896} &
  \textbf{0.968} &
  \textbf{0.981} &
  0.582 &
  0.596 &
  \textbf{0.246} &
  \textbf{0.269} &
  0.273 &
  0.274 \\
+ 5 langs &
  \cellcolor[HTML]{359CE3}0.610 &
  \cellcolor[HTML]{359CE3}0.758 &
  \cellcolor[HTML]{68B5D5}0.773 &
  \cellcolor[HTML]{5EB0D8}0.514 &
  \cellcolor[HTML]{359CE3}0.505 &
  \cellcolor[HTML]{53ABDB}\textbf{0.134} &
  \cellcolor[HTML]{53ABDB}\textbf{0.135} &
  \cellcolor[HTML]{52AADB}0.164 &
  \cellcolor[HTML]{57ADDA}0.202 \\
+ 5 langs + noise &
  \cellcolor[HTML]{6FB9D4}0.701 &
  \cellcolor[HTML]{A6D4C5}0.898 &
  \cellcolor[HTML]{8CC7CC}\textbf{0.831} &
  \cellcolor[HTML]{5CAFD9}0.513 &
  \cellcolor[HTML]{4DA8DD}0.521 &
  \cellcolor[HTML]{7FC1CF}\textbf{0.178} &
  \cellcolor[HTML]{7CBFD0}\textbf{0.184} &
  \cellcolor[HTML]{54ABDB}0.166 &
  \cellcolor[HTML]{7DC0D0}0.266 \\
+ 6 langs &
  \cellcolor[HTML]{C3E3BD}0.831 &
  \cellcolor[HTML]{EDF8B1}\textbf{0.985} &
  \cellcolor[HTML]{EBF7B2}\textbf{0.984} &
  \cellcolor[HTML]{C3E3BD}0.583 &
  \cellcolor[HTML]{CBE7BB}0.605 &
  \cellcolor[HTML]{D3EBB8}\textbf{0.261} &
  \cellcolor[HTML]{D1EAB9}\textbf{0.284} &
  \cellcolor[HTML]{C2E2BD}0.304 &
  \cellcolor[HTML]{A4D3C5}0.331 \\
+ 6 langs + noise &
  \cellcolor[HTML]{DCEFB6}0.870 &
  \cellcolor[HTML]{EBF7B2}\textbf{0.983} &
  \cellcolor[HTML]{EAF6B2}\textbf{0.983} &
  \cellcolor[HTML]{BDE0BE}0.579 &
  \cellcolor[HTML]{B6DCC0}0.591 &
  \cellcolor[HTML]{C9E6BB}\textbf{0.251} &
  \cellcolor[HTML]{C4E3BD}\textbf{0.269} &
  \cellcolor[HTML]{B2DAC1}0.284 &
  \cellcolor[HTML]{9FD1C7}0.323

\end{tabularx}
\caption{Results for systems with continued pre-training only on Swiss German (+ gsw), on 5 other languages (+ 5 langs) and the same languages including Swiss German (+ 6 langs). Darker shades indicate lower scores.
Bold denotes statistically significant improvement compared to their respective baselines COMET-REF or COMET-QE-KIWI. There is no information about significance for tie-optim. accuracy (columns 4-5) and success rate (columns 8-9).
Note that BE and ZH represent the abbreviations for the two Swiss German (GSW) dialect regions under consideration.}
\label{tab:mixed}
%\end{center}
\end{table*}

\section{Correlations on WMT Benchmarks}
\label{app:wmt_correlations}
As discussed in the previous section, we evaluate the performance of our metrics on an official WMT benchmark to monitor their performance on language pairs that do not involve Swiss German. To do this, we reproduce the evaluations from the WMT 2022 metrics task \citep{freitag-etal-2022-results} for a subset of language pairs. We evaluate on the following five language pairs:

\begin{itemize}
    \item \textbf{en-de:} evaluation against MQM ratings collected specifically for the metrics shared task.
    \item \textbf{en-zh:} evaluation against MQM ratings collected specifically for the metrics shared task.
    \item \textbf{de-en:} evaluation against reference-based DA scores collected for the translation shared task.
    \item \textbf{cs-uk:} evaluation against DA + SQM scores collected for the translation shared task.
    \item \textbf{en-liv:} evaluation against DA + SQM scores collected for the translation shared task.
\end{itemize} 

Note that all these languages except for Livonian (liv) are part of the CC-100 corpus\footnote{\url{https://data.statmt.org/cc-100/}} \citep{wenzek-etal-2020-ccnet}. Consequently, they form a part of the training dataset for XLM-R and are thus included in the COMET models. Moreover, English (en), German (de), and Chinese (zh) were incorporated into the mixed continued pre-training, as explained in Section \ref{sec:app-mix}. Lastly, all the languages mentioned above, with the exception of Ukrainian (uk) and Livonian (liv; a language of Latvia), are included in the COMET training data.

This evaluation allows us to assess the effects of our modifications both on language pairs that were included during COMET training, during continued LM pre-training, and those that were not.

The results are shown in the following Tables: 
\ref{tab:sys_corr} (system-level Pearson correlation), \ref{tab:seg_acc} (segment-level accuracy), and \ref{tab:seg_ken} (segment-level Kendall).
We do not report pairwise accuracy here because they cannot be directly compared with the WMT22 results, given that we have only included a subset of the language pairs.
Versions of COMET-ref that were continued pretrained on Swiss German data demonstrate comparable or improved performance compared to the baseline metrics. In contrast, continued pretrained COMET-qe performs worse.
When examining individual languages, we observe that fine-tuning is advantageous for translations into Livonian (liv), which is the only language in our selection not included in XLMR. Conversely, for translations into English, continued pretrained systems, particularly COMET-qe, tend to perform slightly worse.

\begin{table*}[]
\centering
\begin{tabular}{@{\extracolsep{3pt}}rccccc}
\textbf{sys-level Pearson} &  & \textbf{} & \textbf{} & \textbf{} & \textbf{} \\
\textbf{correlation} & de-en & en-de & en-zh & en-liv & cs-uk \\
BLEU &
  \cellcolor[HTML]{359CE3}0.353 &
  \cellcolor[HTML]{359CE3}0.178 &
  \cellcolor[HTML]{359CE3}0.065 &
  \cellcolor[HTML]{359CE3}-0.575 &
  \cellcolor[HTML]{4DA8DD}0.890 \\
chrF++ &
  \cellcolor[HTML]{3A9EE2}0.356 &
  \cellcolor[HTML]{52AADB}0.304 &
  \cellcolor[HTML]{59AEDA}0.203 &
  \cellcolor[HTML]{3C9FE2}-0.517 &
  \cellcolor[HTML]{83C3CE}0.925 \\
COMET-20 &
  \cellcolor[HTML]{BBDFBF}0.424 &
  \cellcolor[HTML]{DBEFB6}0.876 &
  \cellcolor[HTML]{E6F4B3}0.744 &
  \cellcolor[HTML]{EAF6B2}0.893 &
  \cellcolor[HTML]{E0F1B5}0.985 \\
COMET-22 &
  \cellcolor[HTML]{EDF8B1}0.450 &
  \cellcolor[HTML]{DAEEB7}0.873 &
  \cellcolor[HTML]{EAF6B2}0.756 &
  \cellcolor[HTML]{3C9FE2}-0.517 &
  \cellcolor[HTML]{E6F4B3}0.989 \\
COMET-20-QE &
  \cellcolor[HTML]{DFF1B5}0.443 &
  \cellcolor[HTML]{93CBCA}0.577 &
  \cellcolor[HTML]{E9F6B3}0.752 &
  \cellcolor[HTML]{C1E2BD}0.564 &
  \cellcolor[HTML]{AFD9C2}0.953 \\
COMET-KIWI &
  \cellcolor[HTML]{B5DCC0}0.421 &
  \cellcolor[HTML]{BCDFBF}0.748 &
  \cellcolor[HTML]{EDF8B1}0.767 &
  \cellcolor[HTML]{369CE3}-0.563 &
  \cellcolor[HTML]{E3F3B4}0.987 \\ \addlinespace \hline \addlinespace
COMET-ref &
  \cellcolor[HTML]{B9DEBF}0.423 &
  \cellcolor[HTML]{DEF0B6}0.888 &
  \cellcolor[HTML]{C8E5BC}0.626 &
  \cellcolor[HTML]{ECF7B2}0.909 &
  \cellcolor[HTML]{EBF7B2}0.992 \\
+ noise &
  \cellcolor[HTML]{B4DBC1}0.420 &
  \cellcolor[HTML]{E8F5B3}\bf{0.931} &
  \cellcolor[HTML]{C5E4BC}0.618 &
  \cellcolor[HTML]{EDF8B1}0.912 &
  \cellcolor[HTML]{E9F6B2}0.991 \\
+ gsw &
  \cellcolor[HTML]{A1D2C6}0.410 &
  \cellcolor[HTML]{E1F2B5}0.904 &
  \cellcolor[HTML]{99CEC8}0.450 &
  \cellcolor[HTML]{D1EAB9}0.693 &
  \cellcolor[HTML]{DDF0B6}0.983 \\
+ gsw + noise &
  \cellcolor[HTML]{9BCFC8}0.407 &
  \cellcolor[HTML]{E8F5B3}0.930 &
  \cellcolor[HTML]{86C4CD}\underline{0.375} &
  \cellcolor[HTML]{C7E5BC}0.610 &
  \cellcolor[HTML]{C0E1BE}\underline{0.964} \\
+ 5 langs &
  \cellcolor[HTML]{A4D3C5}0.412 &
  \cellcolor[HTML]{E0F1B5}0.897 &
  \cellcolor[HTML]{CFE9B9}0.656 &
  \cellcolor[HTML]{E2F2B4}0.826 &
  \cellcolor[HTML]{EDF8B1}0.993 \\
+ 5 langs + noise &
  \cellcolor[HTML]{AAD6C4}0.415 &
  \cellcolor[HTML]{E8F5B3}\bf{0.933} &
  \cellcolor[HTML]{D0E9B9}0.658 &
  \cellcolor[HTML]{D1EAB9}0.689 &
  \cellcolor[HTML]{E9F6B2}0.991 \\
+ 6 langs &
  \cellcolor[HTML]{AED8C3}0.417 &
  \cellcolor[HTML]{E2F2B4}0.908 &
  \cellcolor[HTML]{CAE6BB}0.636 &
  \cellcolor[HTML]{EAF6B2}0.892 &
  \cellcolor[HTML]{EBF7B2}0.992 \\
+ 6 langs + noise &
  \cellcolor[HTML]{A6D4C5}0.413 &
  \cellcolor[HTML]{EDF8B1}\bf{0.951} &
  \cellcolor[HTML]{C8E5BC}0.626 &
  \cellcolor[HTML]{C9E6BB}0.627 &
  \cellcolor[HTML]{E6F4B3}0.989 \\ \addlinespace
COMET-qe &
  \cellcolor[HTML]{6FB9D4}0.384 &
  \cellcolor[HTML]{76BCD2}0.453 &
  \cellcolor[HTML]{CBE7BB}0.639 &
  \cellcolor[HTML]{C6E4BC}0.598 &
  \cellcolor[HTML]{B0D9C2}0.954 \\
+ noise &
  \cellcolor[HTML]{8AC6CC}0.398 &
  \cellcolor[HTML]{79BED1}0.464 &
  \cellcolor[HTML]{D0E9B9}0.659 &
  \cellcolor[HTML]{C5E4BC}0.589 &
  \cellcolor[HTML]{BBDFBF}0.961 \\
+ gsw &
  \cellcolor[HTML]{4BA7DD}0.365 &
  \cellcolor[HTML]{52AADC}0.300 &
  \cellcolor[HTML]{98CDC9}0.444 &
  \cellcolor[HTML]{DFF1B5}0.806 &
  \cellcolor[HTML]{359CE3}\underline{0.874} \\
+ gsw + noise &
  \cellcolor[HTML]{75BCD2}0.387 &
  \cellcolor[HTML]{5EB0D8}0.354 &
  \cellcolor[HTML]{98CDC8}0.446 &
  \cellcolor[HTML]{E6F4B3}0.859 &
  \cellcolor[HTML]{52AADC}\underline{0.893} \\
+ 5 langs &
  \cellcolor[HTML]{57ADDA}0.371 &
  \cellcolor[HTML]{71BAD3}0.434 &
  \cellcolor[HTML]{CEE8BA}0.650 &
  \cellcolor[HTML]{C8E5BB}0.621 &
  \cellcolor[HTML]{80C1CF}\underline{0.923} \\
+ 5 langs + noise &
  \cellcolor[HTML]{62B2D7}0.377 &
  \cellcolor[HTML]{70B9D3}0.429 &
  \cellcolor[HTML]{D2EAB9}0.667 &
  \cellcolor[HTML]{CBE7BB}0.639 &
  \cellcolor[HTML]{99CEC8}0.939 \\
+ 6 langs &
  \cellcolor[HTML]{59AEDA}0.372 &
  \cellcolor[HTML]{6FB9D4}0.424 &
  \cellcolor[HTML]{D0E9B9}0.657 &
  \cellcolor[HTML]{D2EAB9}0.694 &
  \cellcolor[HTML]{7DC0D0}\underline{0.921} \\
+ 6 langs + noise &
  \cellcolor[HTML]{68B5D6}0.380 &
  \cellcolor[HTML]{73BBD3}0.440 &
  \cellcolor[HTML]{CBE7BB}0.640 &
  \cellcolor[HTML]{D5ECB8}0.725 &
  \cellcolor[HTML]{99CEC8}0.939
\end{tabular}
\caption{System-level Pearson correlation scores for baseline metrics (above) and our trained metrics (below) on a subset of language pairs from the WMT 2022 metrics task. Bold denotes statistically significant improvement compared to their respective baselines COMET-REF or COMET-QE-KIWI, underlined denotes statistically significant decline.}
\label{tab:sys_corr}
\end{table*}

\begin{table*}[]
\centering
\begin{tabular}{@{\extracolsep{3pt}}rccccc}
\textbf{seg-level tie-optim.} &  & \textbf{} & \textbf{} & \textbf{} & \textbf{} \\
\textbf{accuracy} & de-en & en-de & en-zh & en-liv & cs-uk \\
BLEU &
  \cellcolor[HTML]{3DA0E1}0.394 &
  \cellcolor[HTML]{359CE3}0.539 &
  \cellcolor[HTML]{359CE3}0.096 &
  \cellcolor[HTML]{5CAFD9}0.319 &
  \cellcolor[HTML]{60B1D8}0.490 \\
chrF++ &
  \cellcolor[HTML]{359CE3}0.391 &
  \cellcolor[HTML]{4DA8DD}0.545 &
  \cellcolor[HTML]{AED8C2}0.352 &
  \cellcolor[HTML]{359CE3}0.237 &
  \cellcolor[HTML]{359CE3}0.466 \\
COMET-20 &
  \cellcolor[HTML]{C3E3BD}0.439 &
  \cellcolor[HTML]{DCEFB6}0.580 &
  \cellcolor[HTML]{E4F3B4}0.466 &
  \cellcolor[HTML]{E0F1B5}0.589 &
  \cellcolor[HTML]{E5F4B3}0.563 \\
COMET-22 &
  \cellcolor[HTML]{BDE0BE}0.437 &
  \cellcolor[HTML]{EDF8B1}0.584 &
  \cellcolor[HTML]{E5F4B3}0.468 &
  \cellcolor[HTML]{74BBD2}0.368 &
  \cellcolor[HTML]{EDF8B1}0.567 \\
COMET-20-QE &
  \cellcolor[HTML]{CCE7BA}0.442 &
  \cellcolor[HTML]{A3D3C6}0.566 &
  \cellcolor[HTML]{E2F2B4}0.460 &
  \cellcolor[HTML]{BBDFBF}0.513 &
  \cellcolor[HTML]{D8EDB7}0.556 \\
COMET-KIWI &
  \cellcolor[HTML]{73BBD3}0.412 &
  \cellcolor[HTML]{DCEFB6}0.580 &
  \cellcolor[HTML]{E6F4B3}0.470 &
  \cellcolor[HTML]{66B4D6}0.338 &
  \cellcolor[HTML]{EDF8B1}0.567 \\ \addlinespace \hline \addlinespace
COMET-ref &
  \cellcolor[HTML]{C3E3BD}0.439 &
  \cellcolor[HTML]{9FD1C7}0.565 &
  \cellcolor[HTML]{E3F3B4}0.462 &
  \cellcolor[HTML]{C8E5BB}0.540 &
  \cellcolor[HTML]{D8EDB7}0.556 \\
+ noise &
  \cellcolor[HTML]{B4DBC1}0.434 &
  \cellcolor[HTML]{7ABED1}0.556 &
  \cellcolor[HTML]{E1F2B5}0.458 &
  \cellcolor[HTML]{EDF8B1}0.615 &
  \cellcolor[HTML]{E7F5B3}0.564 \\
+ gsw &
  \cellcolor[HTML]{B4DBC1}0.434 &
  \cellcolor[HTML]{66B4D6}0.551 &
  \cellcolor[HTML]{E6F4B3}0.470 &
  \cellcolor[HTML]{B8DDC0}0.507 &
  \cellcolor[HTML]{BFE1BE}0.542 \\
+ gsw + noise &
  \cellcolor[HTML]{AED8C2}0.432 &
  \cellcolor[HTML]{45A4DF}0.543 &
  \cellcolor[HTML]{E6F4B3}0.470 &
  \cellcolor[HTML]{9ED0C7}0.453 &
  \cellcolor[HTML]{ABD7C3}0.531 \\
+ 5 langs &
  \cellcolor[HTML]{D2EAB9}0.444 &
  \cellcolor[HTML]{B3DBC1}0.570 &
  \cellcolor[HTML]{E7F5B3}0.471 &
  \cellcolor[HTML]{E2F2B4}0.593 &
  \cellcolor[HTML]{C1E2BD}0.543 \\
+ 5 langs + noise &
  \cellcolor[HTML]{A2D2C6}0.428 &
  \cellcolor[HTML]{6EB8D4}0.553 &
  \cellcolor[HTML]{EAF6B2}0.478 &
  \cellcolor[HTML]{B5DCC1}0.500 &
  \cellcolor[HTML]{D1EAB9}0.552 \\
+ 6 langs &
  \cellcolor[HTML]{D5ECB8}0.445 &
  \cellcolor[HTML]{A7D5C4}0.567 &
  \cellcolor[HTML]{E9F6B3}0.475 &
  \cellcolor[HTML]{A2D2C6}0.461 &
  \cellcolor[HTML]{CFE9B9}0.551 \\
+ 6 langs + noise &
  \cellcolor[HTML]{A8D5C4}0.430 &
  \cellcolor[HTML]{8AC6CC}0.560 &
  \cellcolor[HTML]{EDF8B1}0.483 &
  \cellcolor[HTML]{C0E1BE}0.523 &
  \cellcolor[HTML]{C4E3BC}0.545 \\ \addlinespace
COMET-qe &
  \cellcolor[HTML]{B1DAC2}0.433 &
  \cellcolor[HTML]{61B2D7}0.550 &
  \cellcolor[HTML]{E6F4B3}0.470 &
  \cellcolor[HTML]{CAE6BB}0.545 &
  \cellcolor[HTML]{D7EDB7}0.555 \\
+ noise &
  \cellcolor[HTML]{BADEBF}0.436 &
  \cellcolor[HTML]{8EC8CB}0.561 &
  \cellcolor[HTML]{E6F4B3}0.470 &
  \cellcolor[HTML]{BEE0BE}0.520 &
  \cellcolor[HTML]{C3E3BD}0.544 \\
+ gsw &
  \cellcolor[HTML]{D5ECB8}0.445 &
  \cellcolor[HTML]{51AADC}0.546 &
  \cellcolor[HTML]{E7F5B3}0.472 &
  \cellcolor[HTML]{DDF0B6}0.583 &
  \cellcolor[HTML]{93CBCA}0.518 \\
+ gsw + noise &
  \cellcolor[HTML]{C9E6BB}0.441 &
  \cellcolor[HTML]{6AB6D5}0.552 &
  \cellcolor[HTML]{E3F3B4}0.463 &
  \cellcolor[HTML]{B7DDC0}0.505 &
  \cellcolor[HTML]{72BAD3}0.500 \\
+ 5 langs &
  \cellcolor[HTML]{D5ECB8}0.445 &
  \cellcolor[HTML]{8EC8CB}0.561 &
  \cellcolor[HTML]{E5F4B4}0.467 &
  \cellcolor[HTML]{BFE1BE}0.522 &
  \cellcolor[HTML]{A9D6C4}0.530 \\
+ 5 langs + noise &
  \cellcolor[HTML]{C3E3BD}0.439 &
  \cellcolor[HTML]{61B2D7}0.550 &
  \cellcolor[HTML]{E3F3B4}0.462 &
  \cellcolor[HTML]{C1E2BD}0.526 &
  \cellcolor[HTML]{97CDC9}0.520 \\
+ 6 langs &
  \cellcolor[HTML]{CFE9BA}0.443 &
  \cellcolor[HTML]{76BCD2}0.555 &
  \cellcolor[HTML]{EBF7B2}0.480 &
  \cellcolor[HTML]{BEE0BE}0.520 &
  \cellcolor[HTML]{A5D4C5}0.528 \\
+ 6 langs + noise &
  \cellcolor[HTML]{EDF8B1}0.453 &
  \cellcolor[HTML]{6AB6D5}0.552 &
  \cellcolor[HTML]{E6F4B3}0.470 &
  \cellcolor[HTML]{BDE0BE}0.517 &
  \cellcolor[HTML]{A2D2C6}0.526
\end{tabular}
\caption{Segment-level accuracy scores (the darker the lower) for baseline metrics (above) and our trained metrics (below) on a subset of language pairs from the WMT 2022 metrics task. There is no information about significance.}
\label{tab:seg_acc}
\end{table*}

\begin{table*}[]
\centering
\begin{tabular}{@{\extracolsep{3pt}}rccccc}
\textbf{seg-level Kendall} &  & \textbf{} & \textbf{} & \textbf{} & \textbf{} \\
\textbf{correlation} & de-en & en-de & en-zh & en-liv & cs-uk \\
BLEU &
  \cellcolor[HTML]{4DA8DD}0.009 &
  \cellcolor[HTML]{4AA6DE}0.169 &
  \cellcolor[HTML]{359CE3}0.032 &
  \cellcolor[HTML]{359CE3}-0.158 &
  \cellcolor[HTML]{5EB0D8}0.133 \\
chrF++ &
  \cellcolor[HTML]{359CE3}0.007 &
  \cellcolor[HTML]{359CE3}0.146 &
  \cellcolor[HTML]{5DB0D8}0.056 &
  \cellcolor[HTML]{359CE3}-0.158 &
  \cellcolor[HTML]{359CE3}0.086 \\
COMET-20 &
  \cellcolor[HTML]{BBDFBF}0.018 &
  \cellcolor[HTML]{D6ECB8}0.319 &
  \cellcolor[HTML]{EDF8B1}0.141 &
  \cellcolor[HTML]{EAF6B2}0.208 &
  \cellcolor[HTML]{DFF1B5}0.280 \\
COMET-22 &
  \cellcolor[HTML]{C8E5BB}0.019 &
  \cellcolor[HTML]{EDF8B1}0.343 &
  \cellcolor[HTML]{E6F4B3}0.137 &
  \cellcolor[HTML]{4CA7DD}-0.111 &
  \cellcolor[HTML]{EDF8B1}0.295 \\
COMET-20-QE &
  \cellcolor[HTML]{EDF8B1}0.022 &
  \cellcolor[HTML]{87C5CD}0.234 &
  \cellcolor[HTML]{CEE8BA}0.123 &
  \cellcolor[HTML]{C1E2BD}0.126 &
  \cellcolor[HTML]{C8E5BB}0.254 \\ 
COMET-KIWI &
  \cellcolor[HTML]{A3D3C5}0.016 &
  \cellcolor[HTML]{84C3CE}0.231 &
  \cellcolor[HTML]{CEE8BA}0.123 &
  \cellcolor[HTML]{3A9EE2}-0.147 &
  \cellcolor[HTML]{E0F1B5}0.281 \\ \addlinespace \hline \addlinespace
COMET-ref &
  \cellcolor[HTML]{97CDC9}0.015 &
  \cellcolor[HTML]{D7EDB7}0.320 &
  \cellcolor[HTML]{E9F6B2}0.139 &
  \cellcolor[HTML]{EDF8B1}0.213 &
  \cellcolor[HTML]{D4EBB8}0.267 \\
+ noise &
  \cellcolor[HTML]{C8E5BB}0.019 &
  \cellcolor[HTML]{CEE8BA}\underline{0.310} &
  \cellcolor[HTML]{D1EAB9}0.125 &
  \cellcolor[HTML]{D5ECB8}\underline{0.165} &
  \cellcolor[HTML]{C6E4BC}\underline{0.251} \\
+ gsw &
  \cellcolor[HTML]{A3D3C5}0.016 &
  \cellcolor[HTML]{BEE0BE}\underline{0.293} &
  \cellcolor[HTML]{C9E6BB}0.120 &
  \cellcolor[HTML]{B2DAC1}\underline{0.096} &
  \cellcolor[HTML]{AFD9C2}\underline{0.225} \\
+ gsw + noise &
  \cellcolor[HTML]{D4EBB8}0.020 &
  \cellcolor[HTML]{C2E2BD}\underline{0.298} &
  \cellcolor[HTML]{A9D6C4}0.101 &
  \cellcolor[HTML]{A0D1C6}\underline{0.059} &
  \cellcolor[HTML]{A4D3C5}\underline{0.213} \\
+ 5 langs &
  \cellcolor[HTML]{AFD9C2}0.017 &
  \cellcolor[HTML]{D3EBB8}\underline{0.316} &
  \cellcolor[HTML]{DCEFB6}0.131 &
  \cellcolor[HTML]{C2E2BD}\underline{0.127} &
  \cellcolor[HTML]{C7E5BC}\underline{0.252} \\
+ 5 langs + noise &
  \cellcolor[HTML]{BBDFBF}0.018 &
  \cellcolor[HTML]{D8EDB7}0.321 &
  \cellcolor[HTML]{D7EDB7}0.128 &
  \cellcolor[HTML]{B2DAC1}\underline{0.095} &
  \cellcolor[HTML]{BADEBF}\underline{0.238} \\
+ 6 langs &
  \cellcolor[HTML]{AFD9C2}0.017 &
  \cellcolor[HTML]{CDE8BA}\underline{0.309} &
  \cellcolor[HTML]{DFF1B5}0.133 &
  \cellcolor[HTML]{C8E5BB}\underline{0.140} &
  \cellcolor[HTML]{C1E2BD}\underline{0.246} \\
+ 6 langs + noise &
  \cellcolor[HTML]{BBDFBF}\bf{0.018} &
  \cellcolor[HTML]{CDE8BA}0.309 &
  \cellcolor[HTML]{D3EBB8}\underline{0.126} &
  \cellcolor[HTML]{A6D4C5}\underline{0.070} &
  \cellcolor[HTML]{B7DDC0}\underline{0.234} \\ \addlinespace
COMET-qe &
  \cellcolor[HTML]{AFD9C2}0.017 &
  \cellcolor[HTML]{7EC0CF}0.225 &
  \cellcolor[HTML]{CBE7BB}0.121 &
  \cellcolor[HTML]{CEE8BA}0.152 &
  \cellcolor[HTML]{B8DDC0}0.235 \\
+ noise &
  \cellcolor[HTML]{8AC6CC}0.014 &
  \cellcolor[HTML]{81C2CF}\bf{0.228} &
  \cellcolor[HTML]{BFE1BE}\underline{0.114} &
  \cellcolor[HTML]{C7E5BC}0.137 &
  \cellcolor[HTML]{A8D5C4}\underline{0.217} \\
+ gsw &
  \cellcolor[HTML]{7EC0CF}0.013 &
  \cellcolor[HTML]{52AADB}\underline{0.178} &
  \cellcolor[HTML]{9BCFC8}\underline{0.093} &
  \cellcolor[HTML]{CBE7BB}0.146 &
  \cellcolor[HTML]{77BDD2}\underline{0.161} \\
+ gsw + noise &
  \cellcolor[HTML]{7EC0CF}0.013 &
  \cellcolor[HTML]{56ACDA}\underline{0.182} &
  \cellcolor[HTML]{9DD0C7}\underline{0.094} &
  \cellcolor[HTML]{B5DCC0}\underline{0.102} &
  \cellcolor[HTML]{77BDD1}0.162 \\
+ 5 langs &
  \cellcolor[HTML]{D4EBB8}0.020 &
  \cellcolor[HTML]{74BBD2}\underline{0.214} &
  \cellcolor[HTML]{C1E2BD}\underline{0.115} &
  \cellcolor[HTML]{CBE7BB}0.145 &
  \cellcolor[HTML]{A5D4C5}\underline{0.214} \\
+ 5 langs + noise &
  \cellcolor[HTML]{C8E5BB}0.019 &
  \cellcolor[HTML]{77BDD1}\underline{0.217} &
  \cellcolor[HTML]{C4E3BD}0.117 &
  \cellcolor[HTML]{C9E6BB}0.142 &
  \cellcolor[HTML]{9CCFC8}\underline{0.203}\\
+ 6 langs &
  \cellcolor[HTML]{97CDC9}0.015 &
  \cellcolor[HTML]{76BCD2}\underline{0.216} &
  \cellcolor[HTML]{C4E3BD}0.117 &
  \cellcolor[HTML]{D6ECB8}0.167 &
  \cellcolor[HTML]{97CDC9}\underline{0.198} \\
+ 6 langs + noise &
  \cellcolor[HTML]{A3D3C5}0.016 &
  \cellcolor[HTML]{72BAD3}\underline{0.212} &
  \cellcolor[HTML]{BFE1BE}\underline{0.114} &
  \cellcolor[HTML]{CCE7BA}0.147 &
  \cellcolor[HTML]{8DC8CC}\underline{0.186}
\end{tabular}
\caption{Segment-level Kendall correlation scores for baseline metrics (above) and our trained metrics (below) on a subset of language pairs from the WMT 2022 metrics task. Bold denotes statistically significant improvement compared to their respective baselines COMET-REF or COMET-QE-KIWI, underlined denotes statistically significant decline.}
\label{tab:seg_ken}
\end{table*}

\section{Pairwise Comparison Plots}
\label{app:plots}

\begin{figure*}%
    \centering
    \subfloat{\includegraphics[scale=0.25]{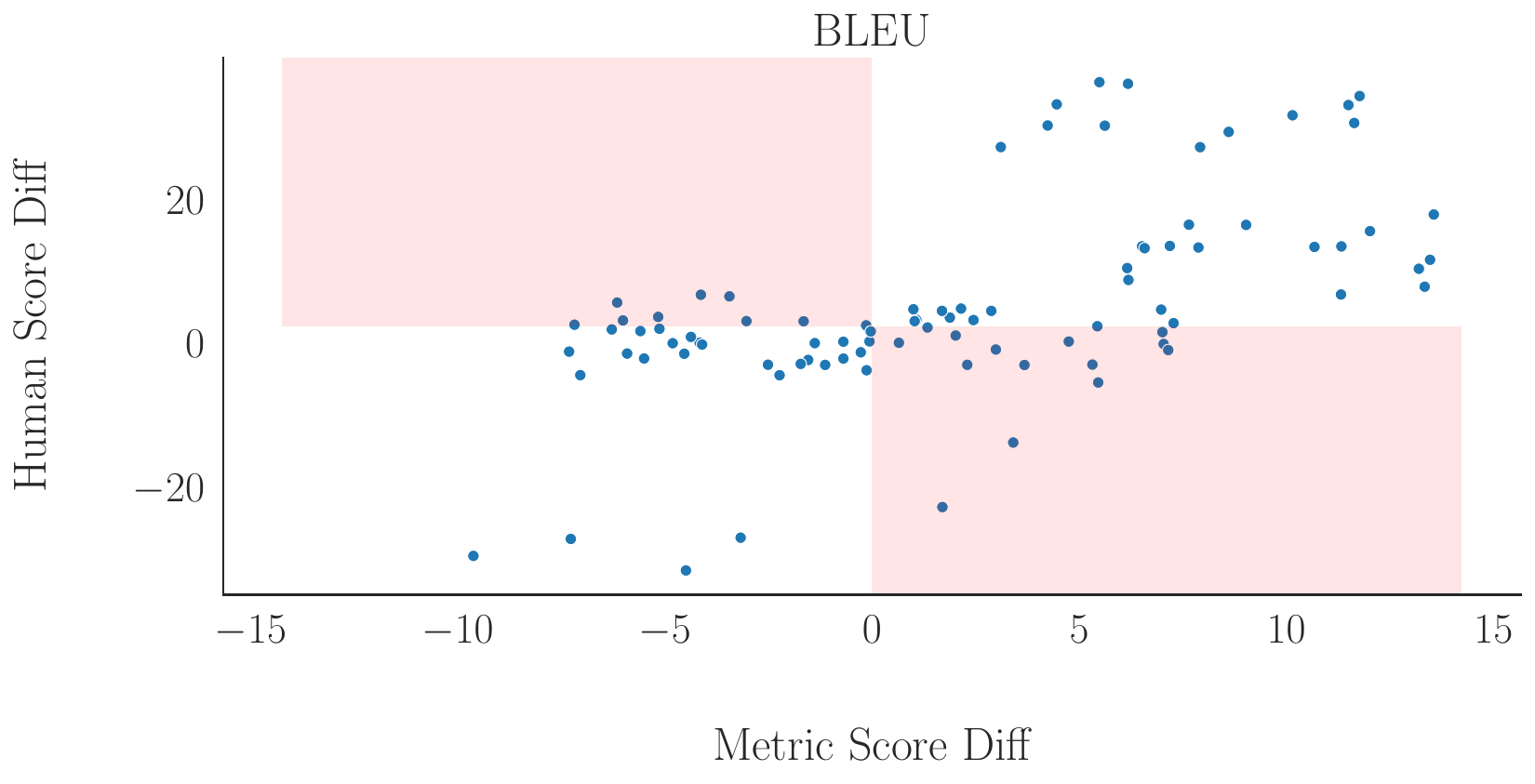}}%
    \qquad
    \subfloat{\includegraphics[scale=0.25]{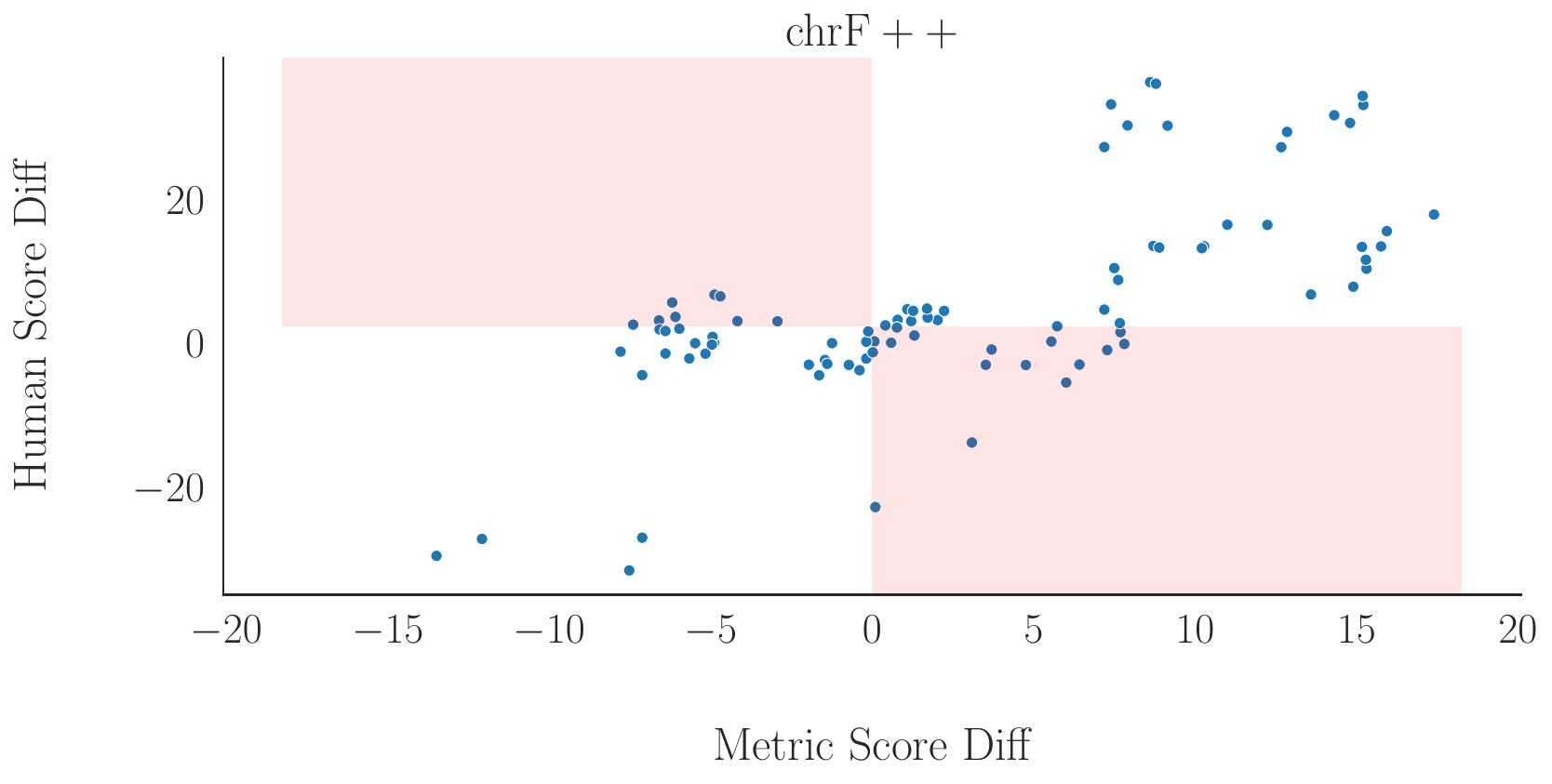}}%

    \subfloat{\includegraphics[scale=0.25]{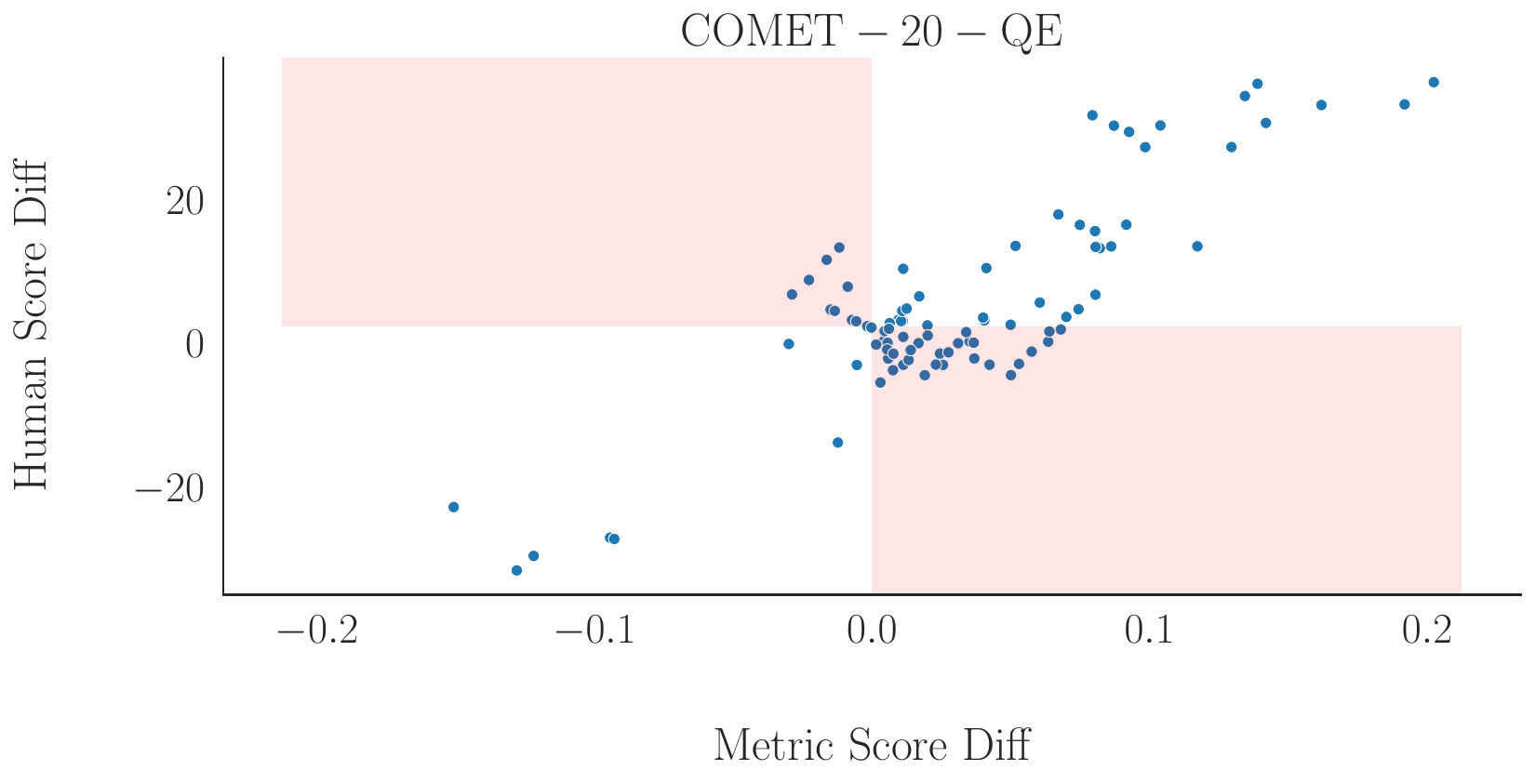}}%
    \qquad
    \subfloat{\includegraphics[scale=0.25]{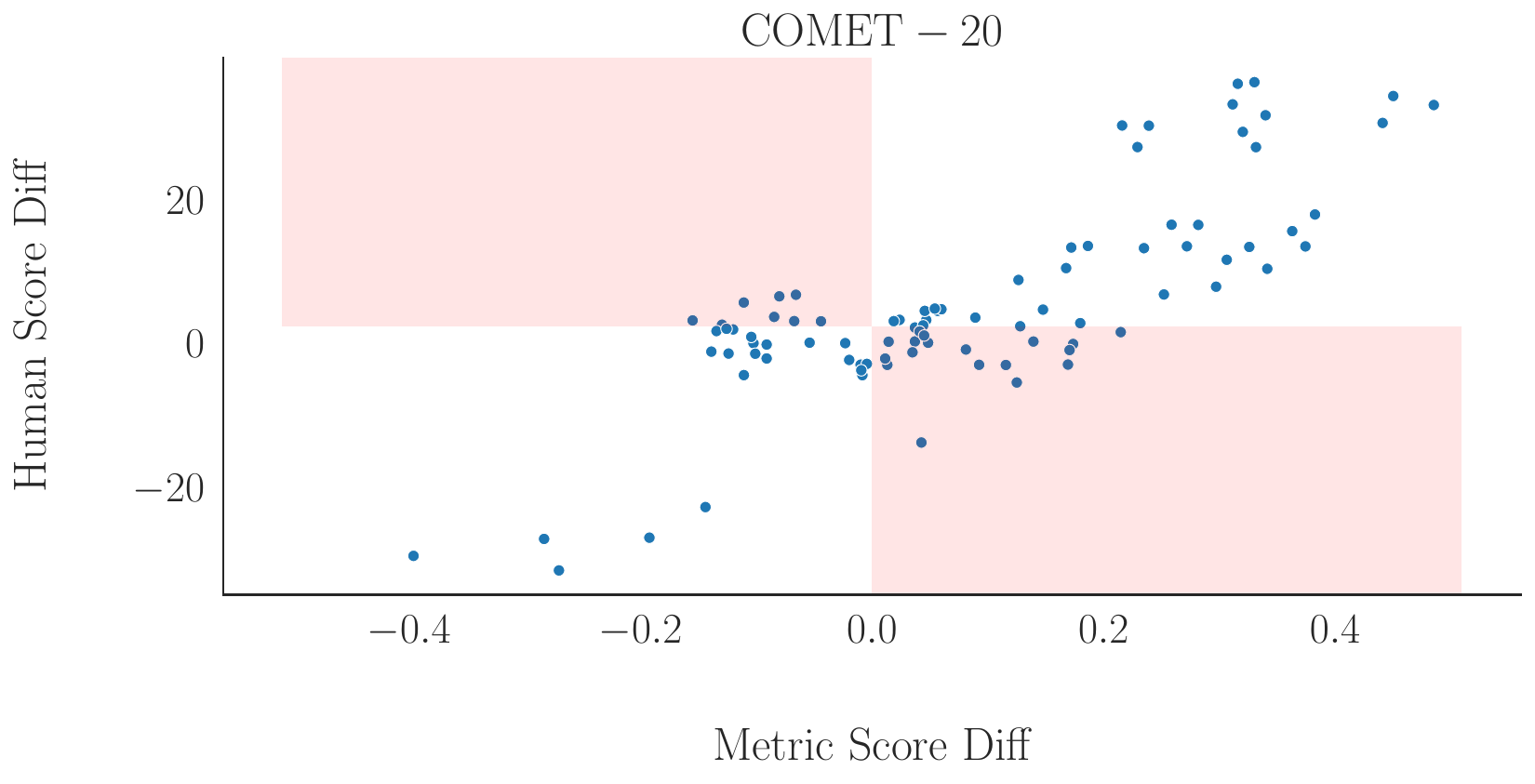}}%

    \subfloat{\includegraphics[scale=0.25]{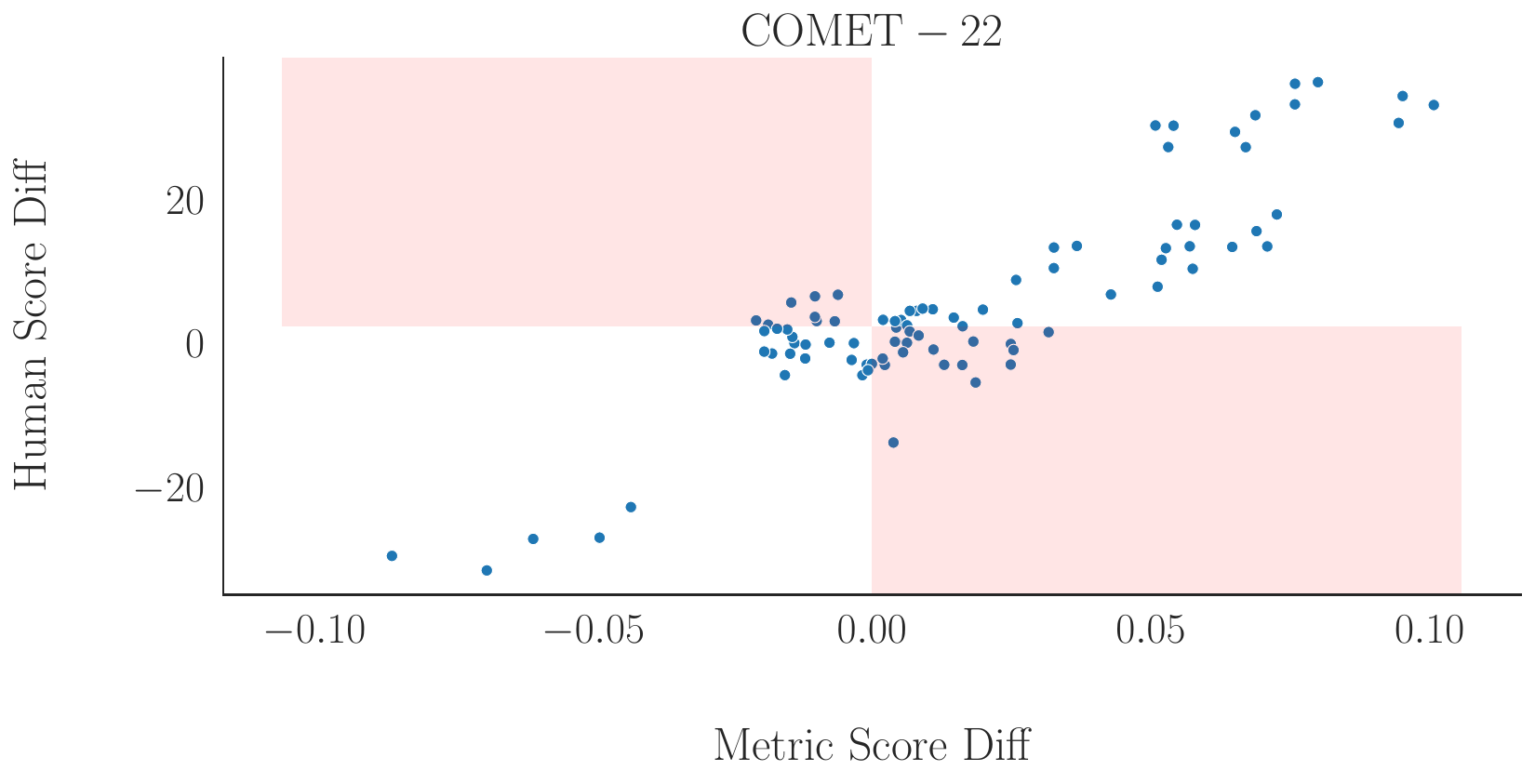}}%
    \qquad
    \subfloat{\includegraphics[scale=0.25]{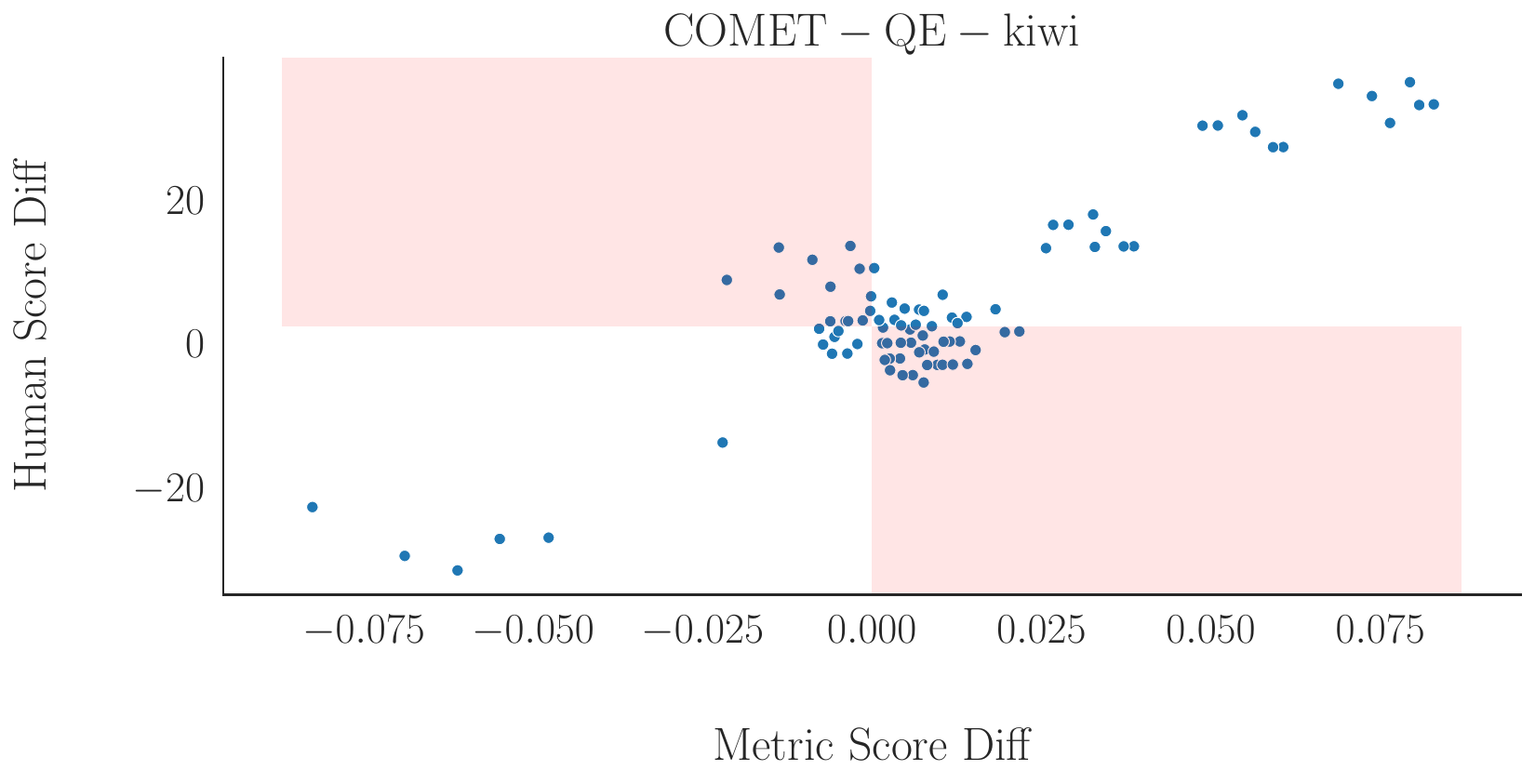}}%
    \caption{Pairwise comparison plots for existing metrics.}%
    \label{fig:plots_baselines}%
\end{figure*}

\begin{figure*}%
    \centering
    \subfloat{\includegraphics[scale=0.25]{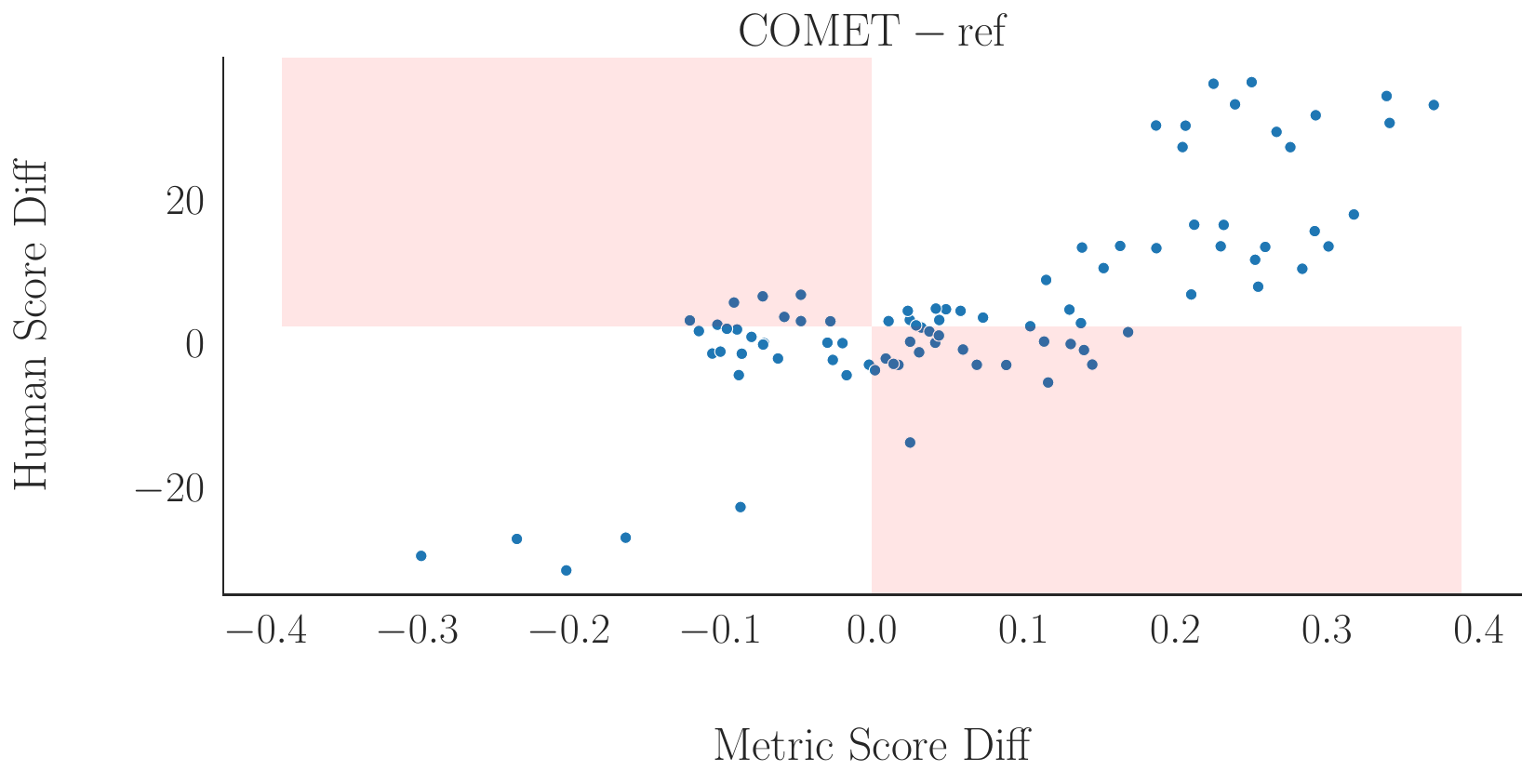}}%
    \qquad
    \subfloat{\includegraphics[scale=0.25]{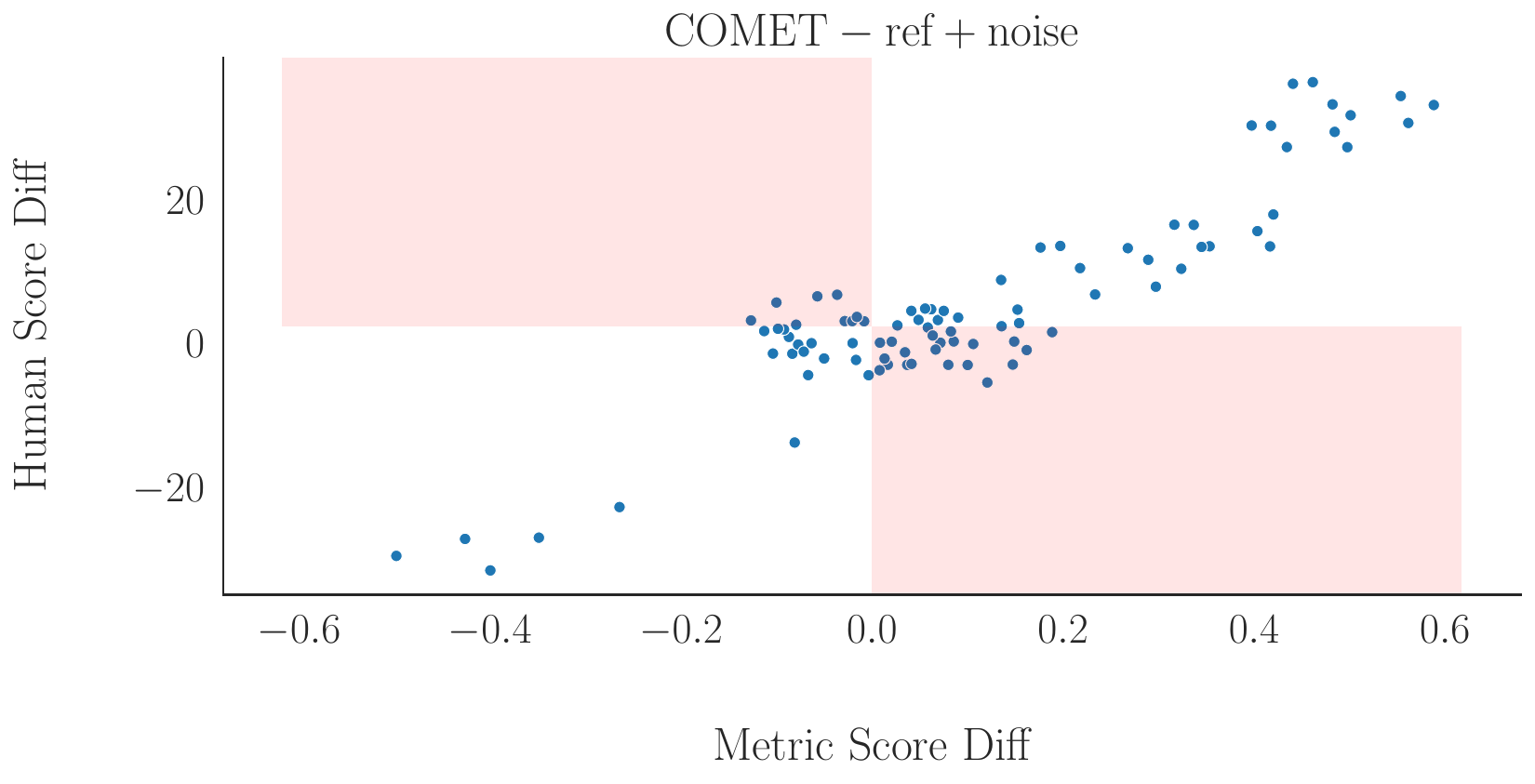}}%

    \subfloat{\includegraphics[scale=0.25]{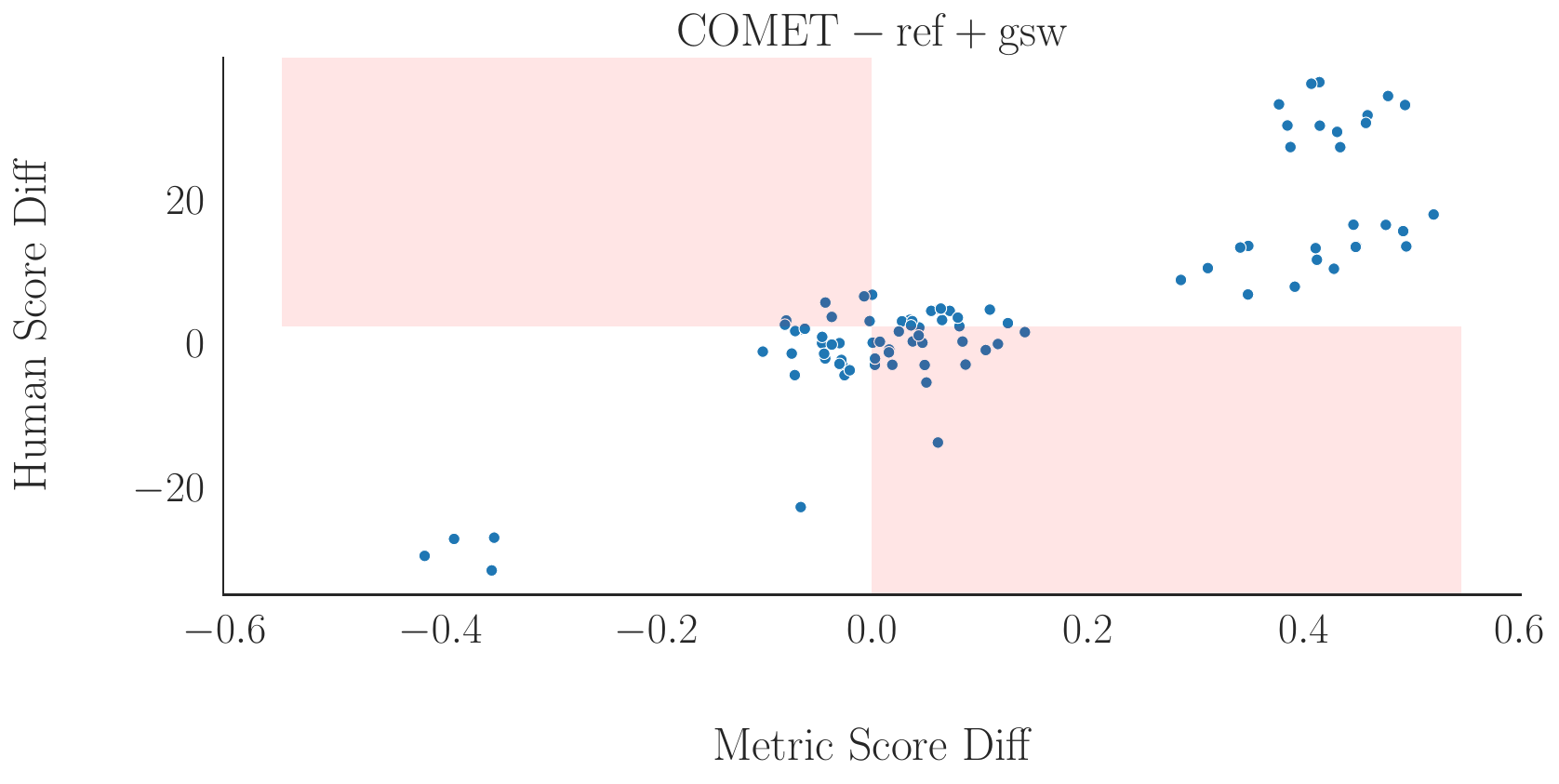}}%
    \qquad
    \subfloat{\includegraphics[scale=0.25]{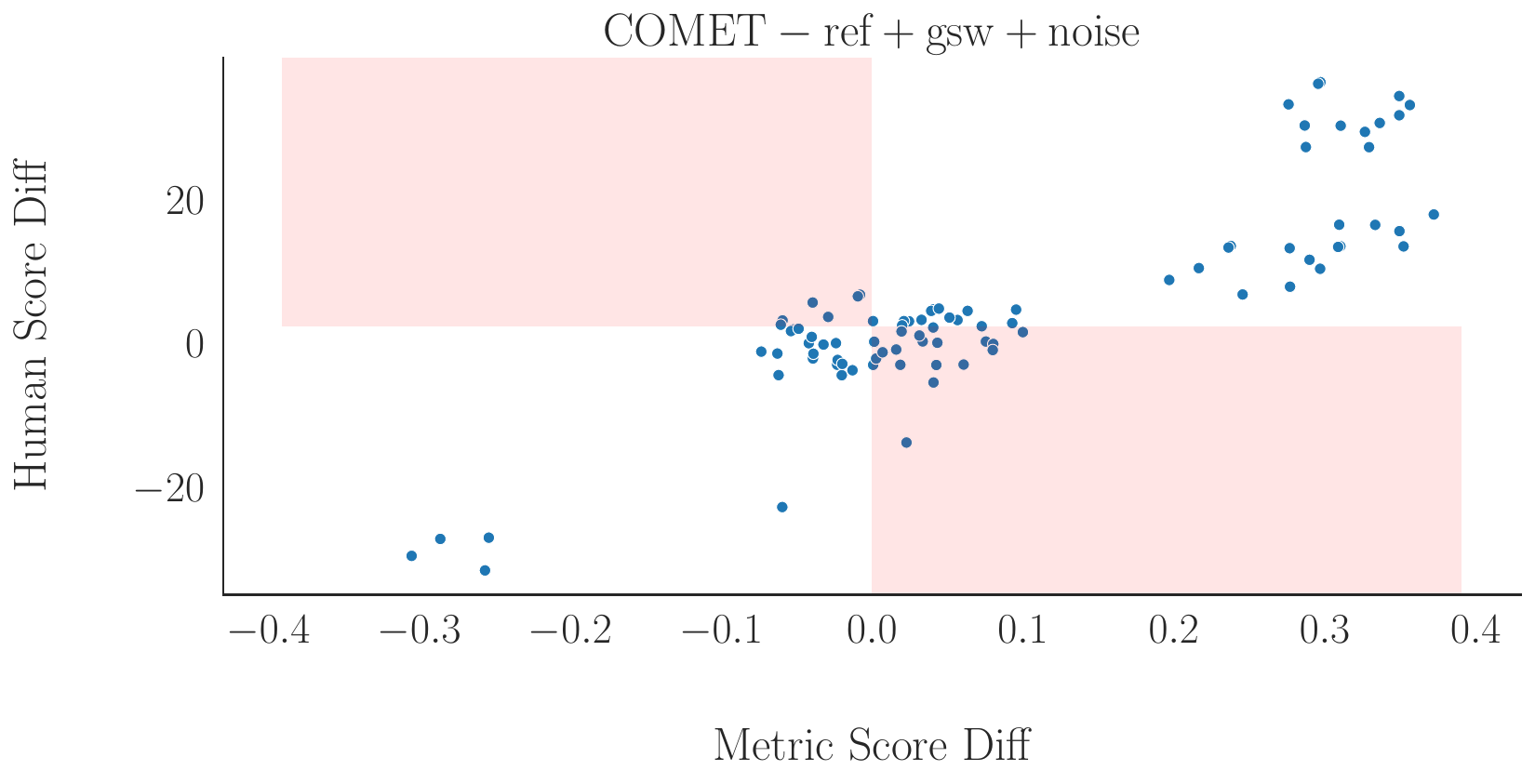}}%

    \subfloat{\includegraphics[scale=0.25]{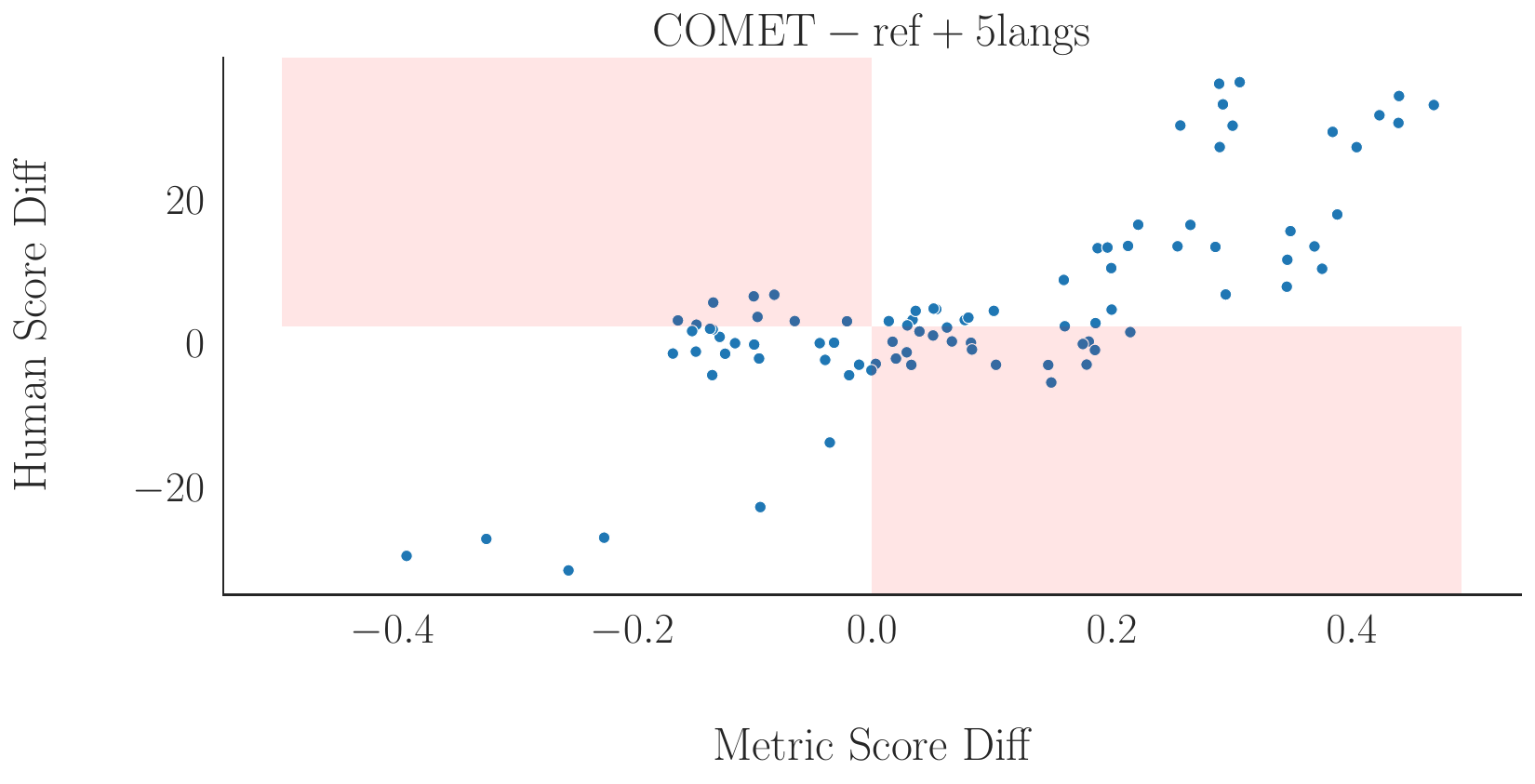}}%
    \qquad
    \subfloat{\includegraphics[scale=0.25]{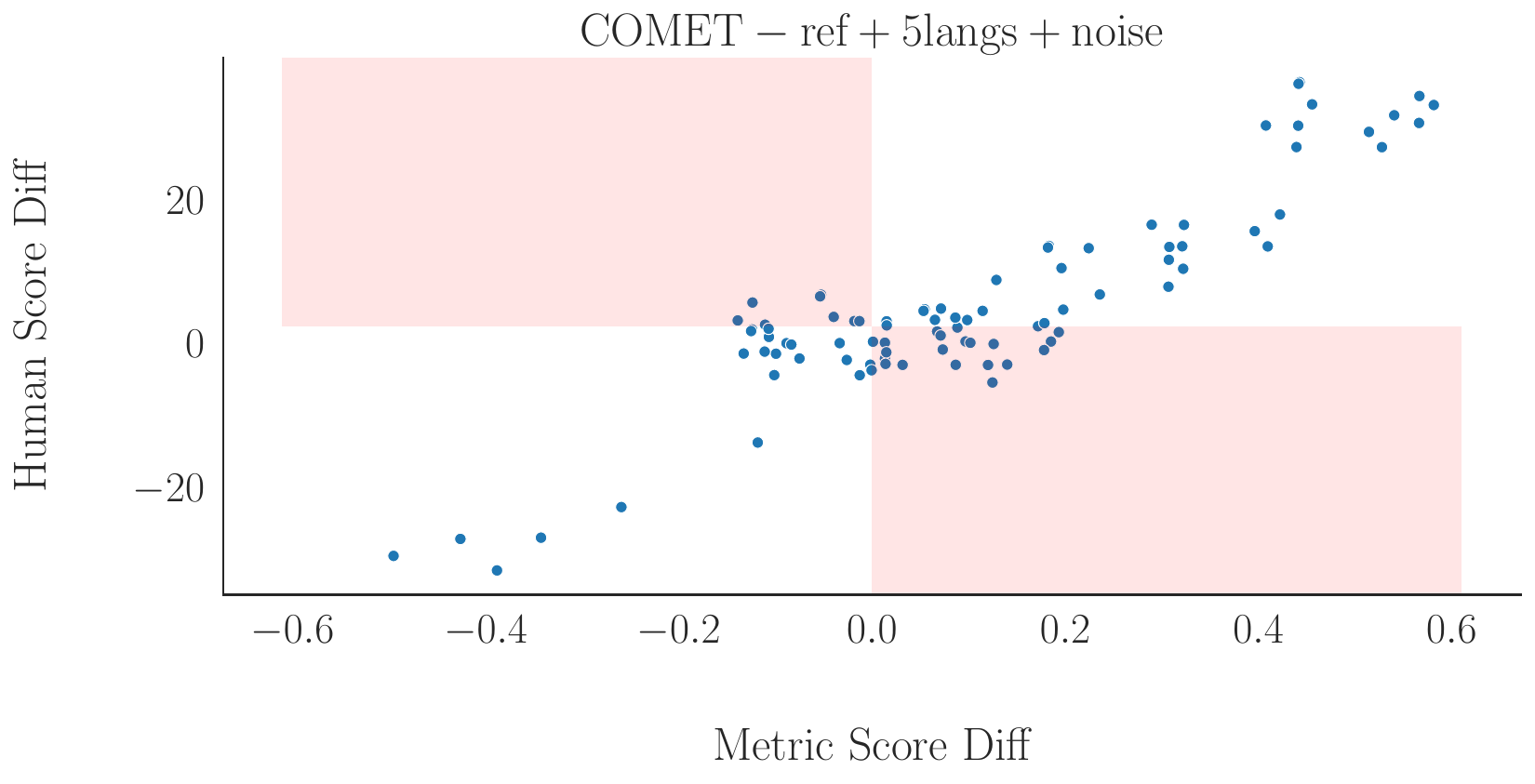}}%
    
    \subfloat{\includegraphics[scale=0.25]{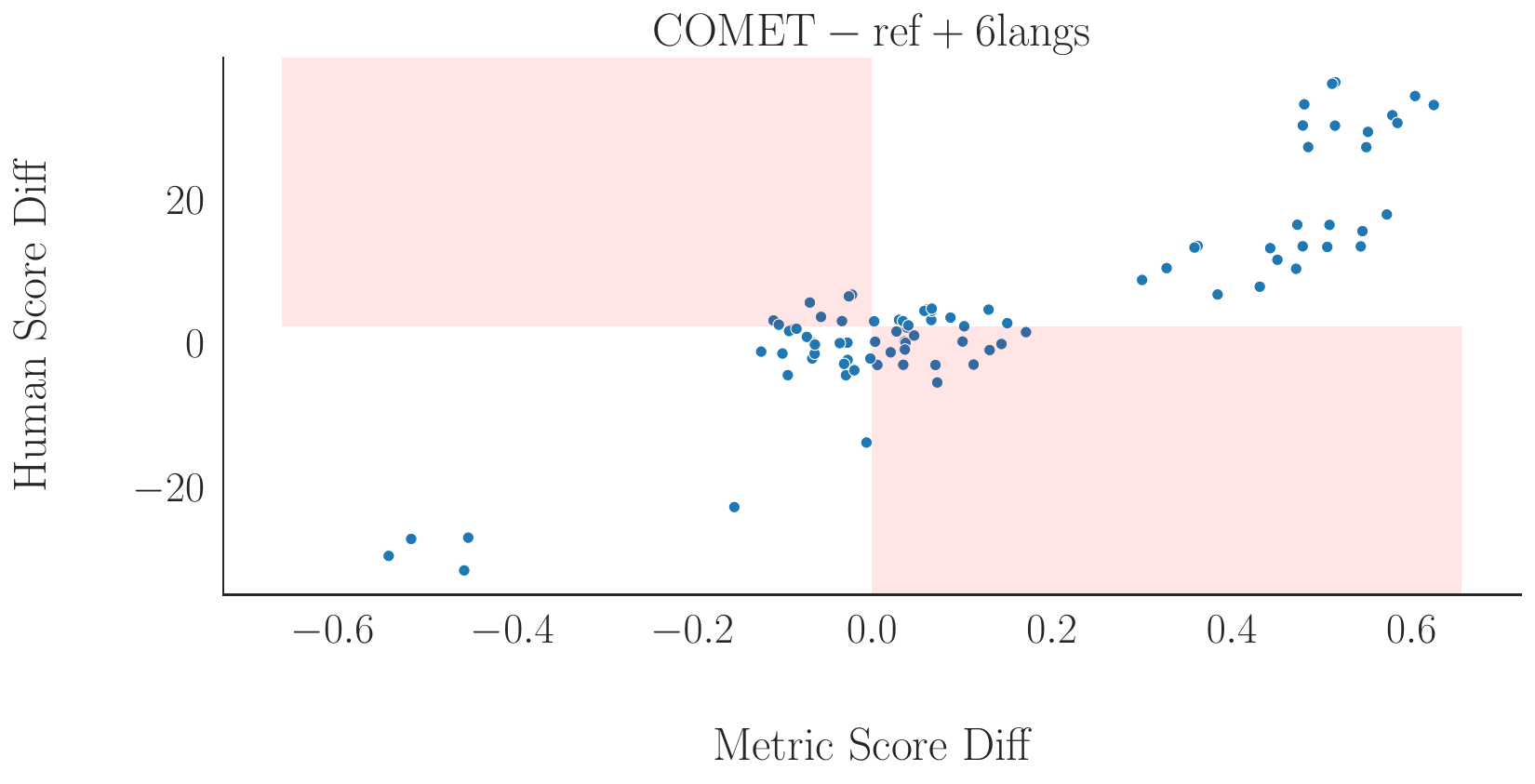}}%
    \qquad
    \subfloat{\includegraphics[scale=0.25]{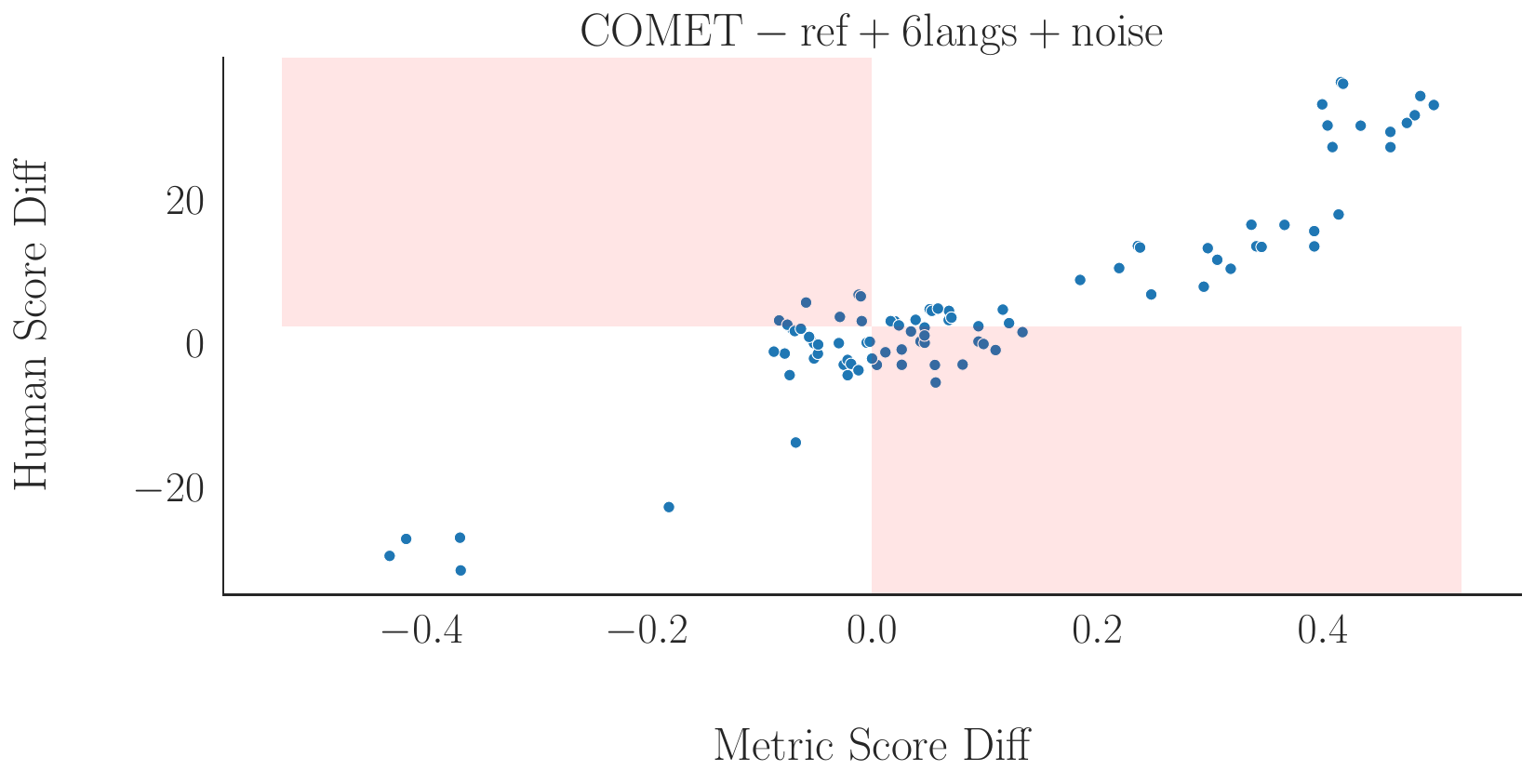}}%
    \caption{Pairwise comparison plots for the COMET-ref metrics trained for this work.}%
    \label{fig:plots_comet-ref}%
\end{figure*}

\begin{figure*}%
    \centering
    \subfloat{\includegraphics[scale=0.25]{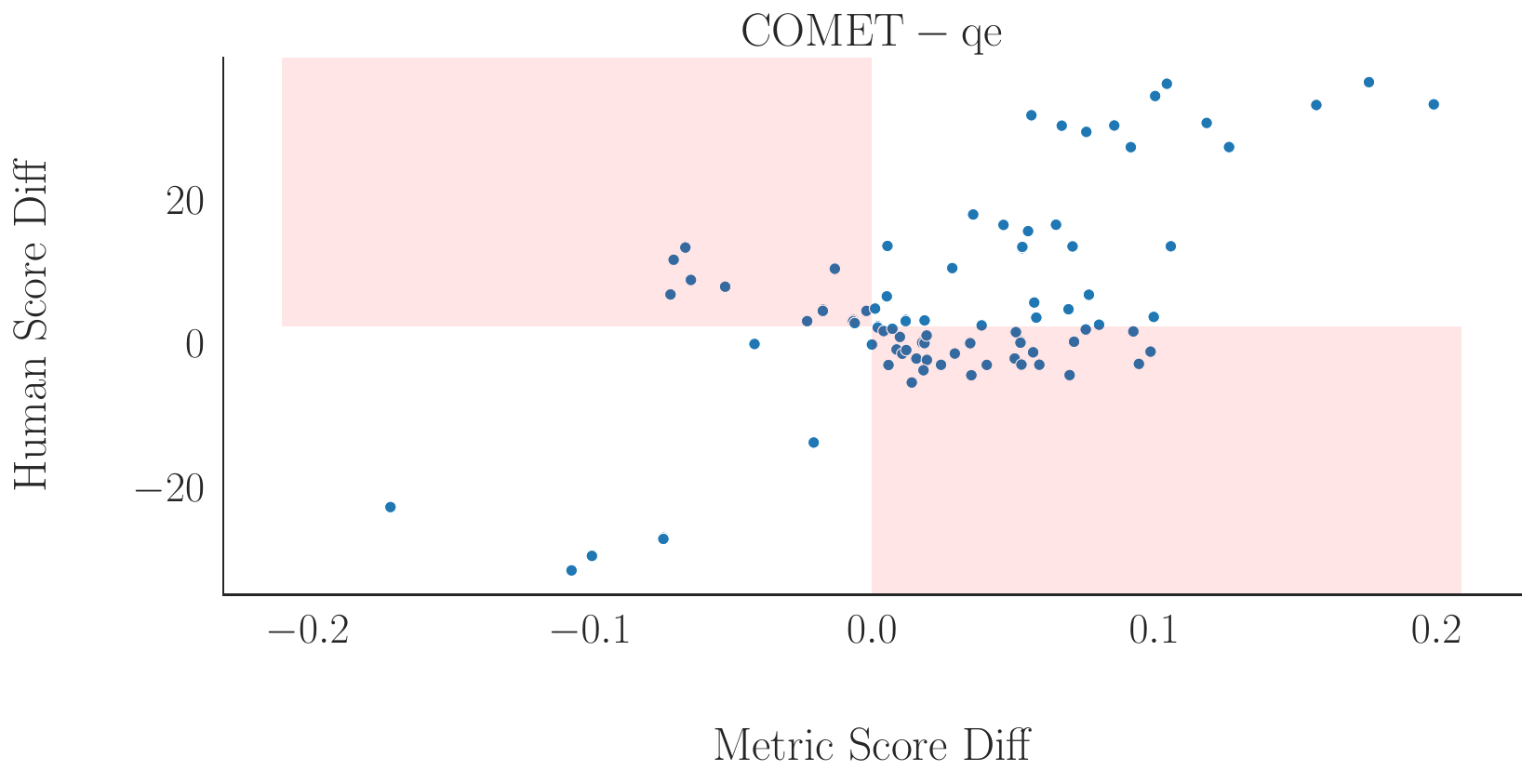}}%
    \qquad
    \subfloat{\includegraphics[scale=0.25]{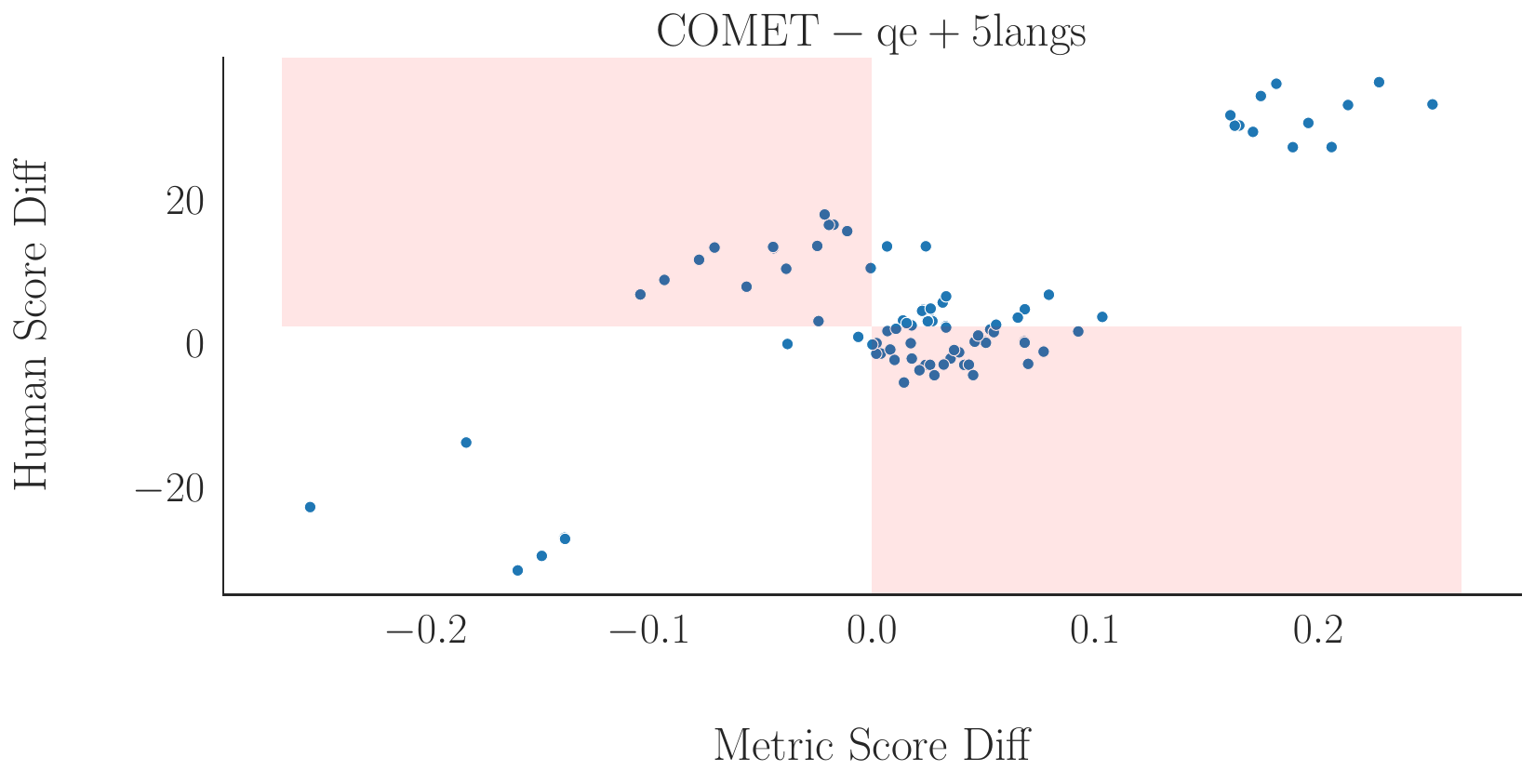}}%

    \subfloat{\includegraphics[scale=0.25]{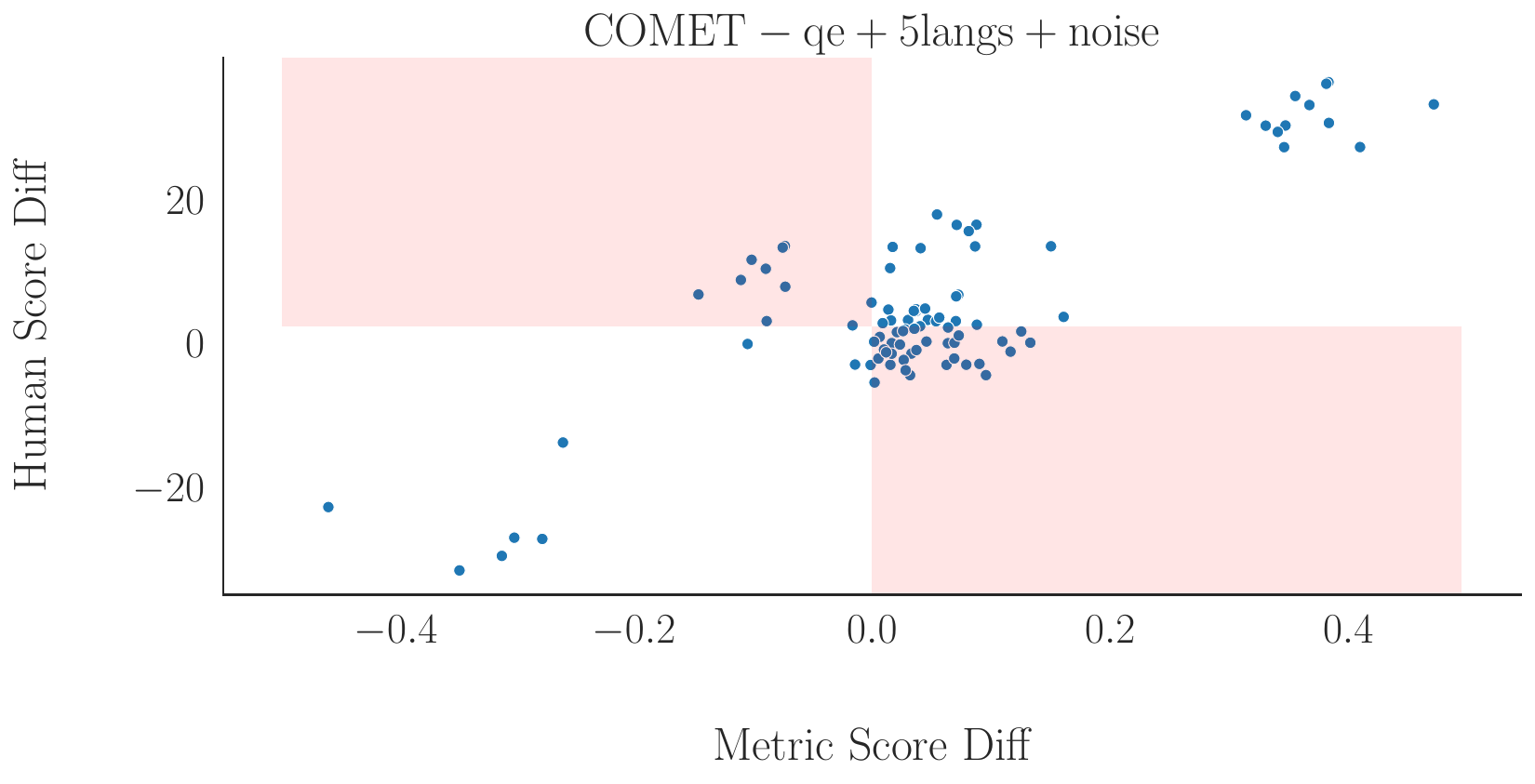}}%
    \qquad
    \subfloat{\includegraphics[scale=0.25]{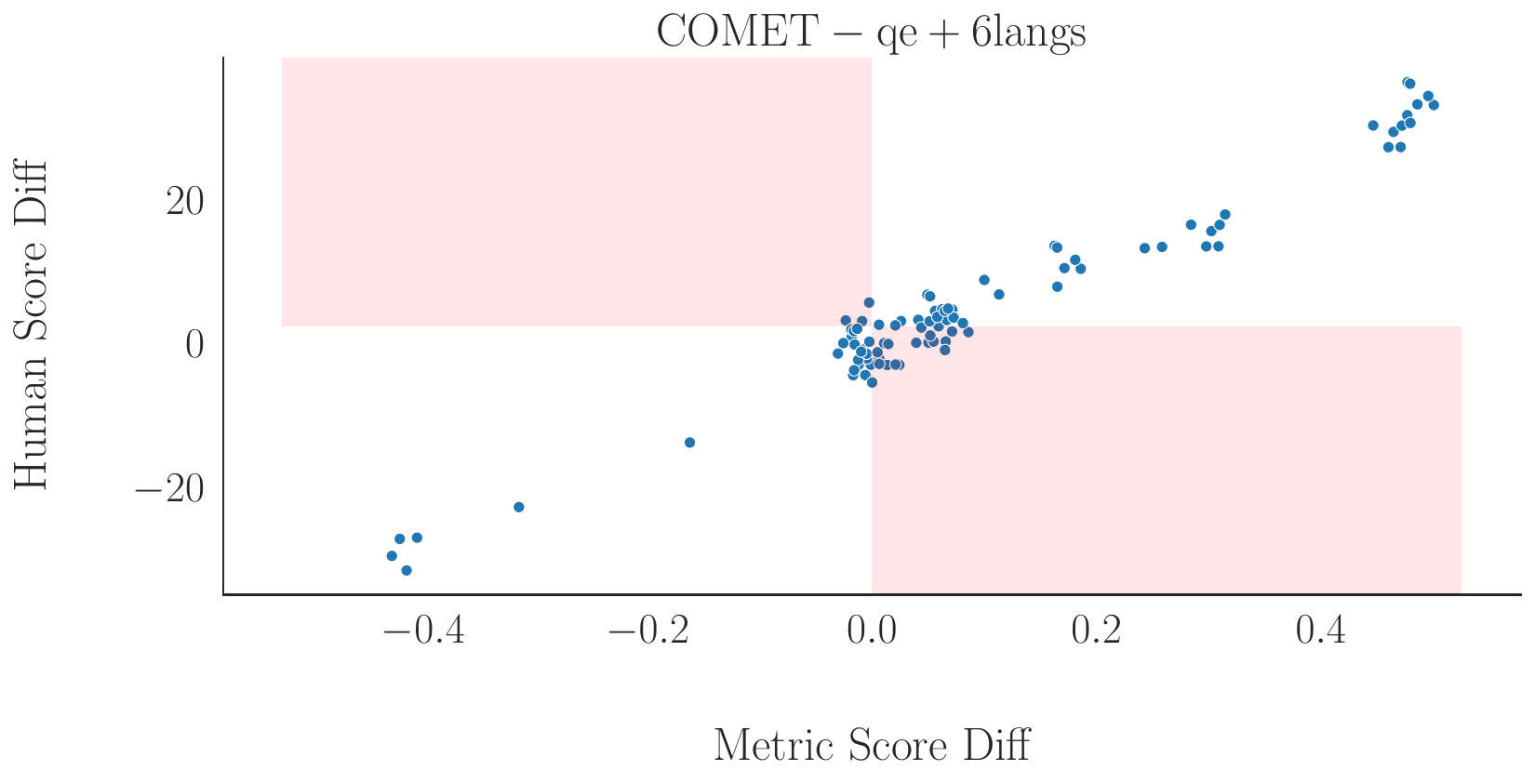}}%

    \subfloat{\includegraphics[scale=0.25]{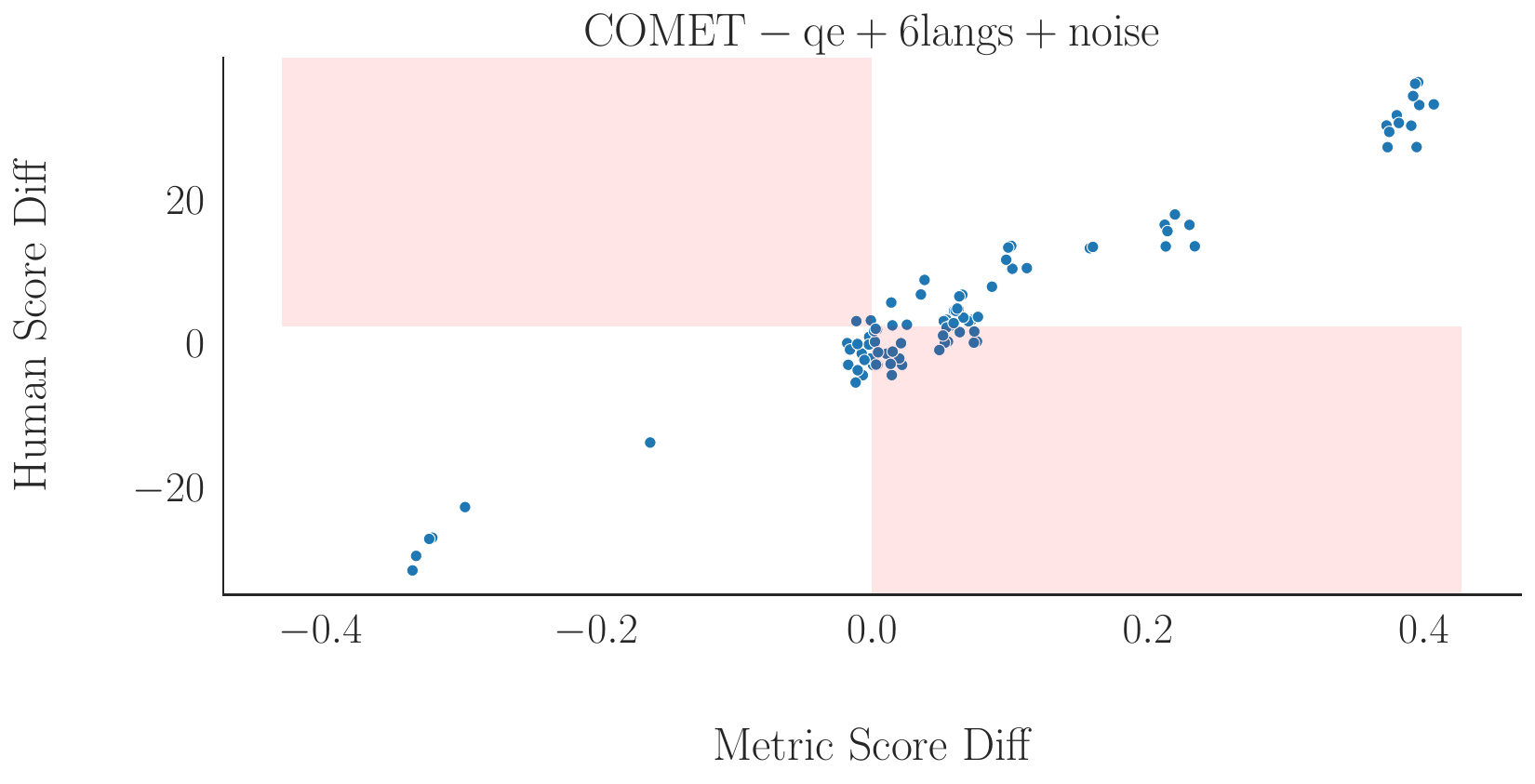}}%
    \qquad
    \subfloat{\includegraphics[scale=0.25]{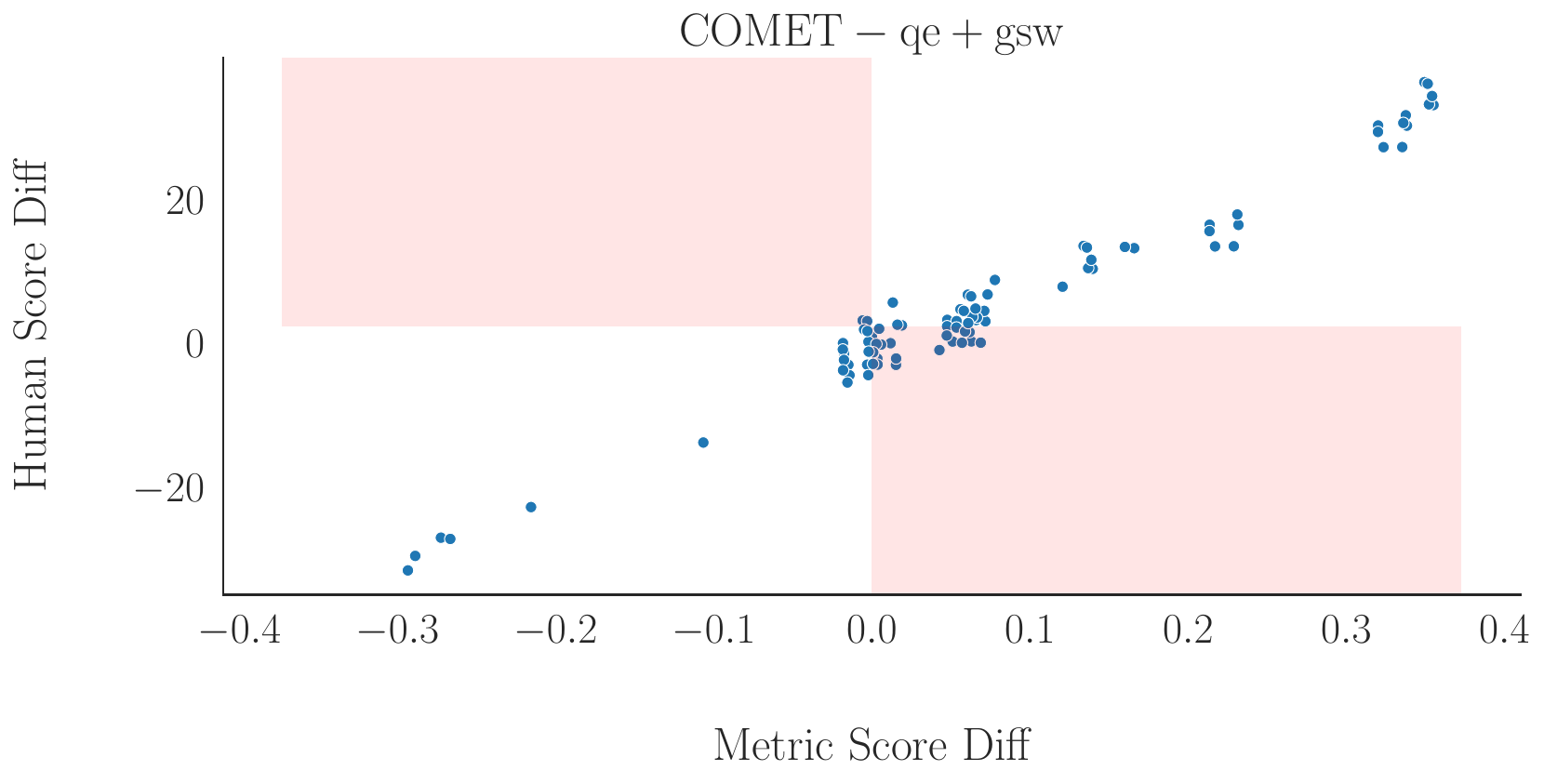}}%

    \subfloat{\includegraphics[scale=0.25]{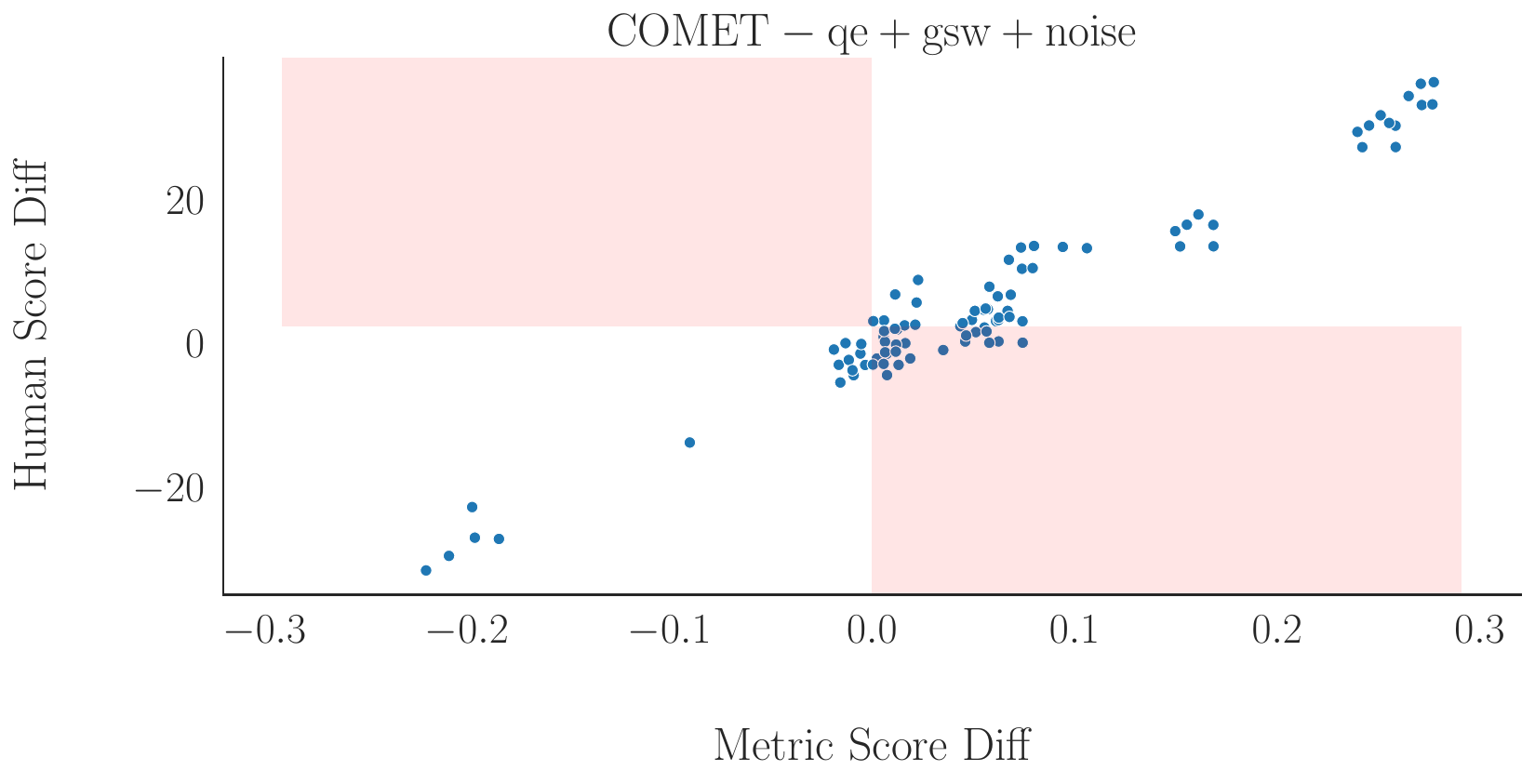}}%
    \qquad
    \subfloat{\includegraphics[scale=0.25]{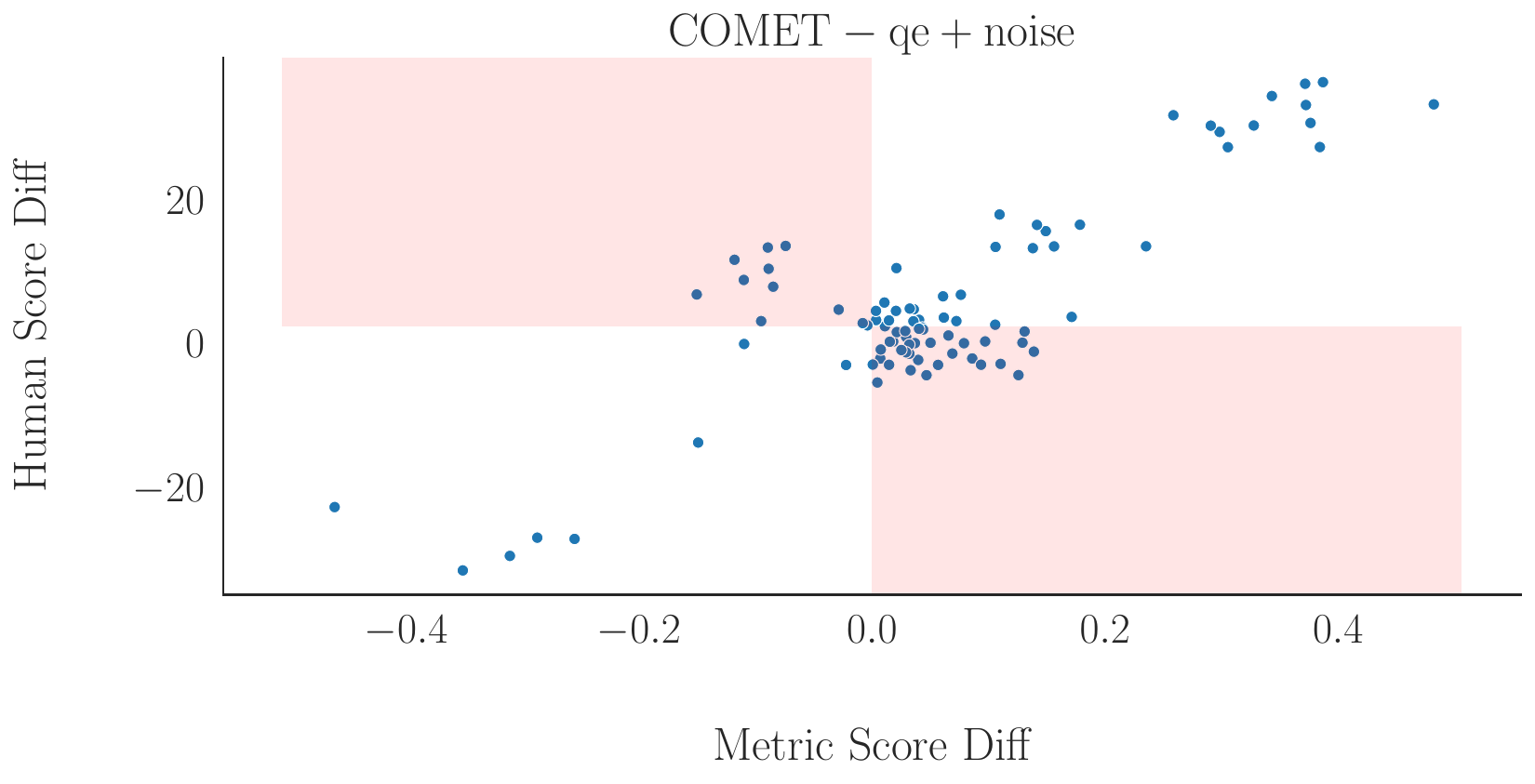}}%
    \caption{Pairwise comparison plots for the COMET-qe metrics trained for this work.}%
    \label{fig:plots_comet-qe}%
\end{figure*}

In the subsequent plots displayed in Figures \ref{fig:plots_baselines} (existing metrics), \ref{fig:plots_comet-ref} (our trained COMET-ref metrics), and \ref{fig:plots_comet-qe} (our trained COMET-qe metrics), every point represents a difference in average human judgment (y-axis) and a difference in automatic metric (x-axis) over a pair of systems. Metrics disagree with human ranking for system pairs in pink quadrants. These plots follow the example of Figure 1 in \cite{kocmi-etal-2021-ship}.

\end{document}